\documentclass[journal]{IEEEtran}
\usepackage{amsmath,amsfonts}
\usepackage{algorithmic}
\usepackage{algorithm}
\usepackage{array}
\usepackage[caption=false,font=normalsize,labelfont=sf,textfont=sf]{subfig}
\usepackage{textcomp}
\usepackage{stfloats}
\usepackage{url}
\usepackage{verbatim}
\usepackage{graphicx}
\usepackage{cite}
\usepackage[table]{xcolor}
\usepackage{colortbl}
\usepackage{multirow}
\usepackage{bbding}
\usepackage{algorithmic}
\usepackage{algorithm}
\usepackage{xspace}
\usepackage{booktabs}
\usepackage{enumitem}

\usepackage{pifont}
\usepackage{color}

\usepackage[pagebackref=false,breaklinks=true,colorlinks,bookmarks=false]{hyperref}

\definecolor{mygray}{gray}{.9}
\definecolor{lightgray}{gray}{.95}
\definecolor{ggray}{RGB}{127,127,127}
\definecolor{reda}{RGB}{192,0,0}
\definecolor{redb}{RGB}{217,148,143}
\definecolor{myyellow}{RGB}{190,144,0}
\definecolor{mygreen}{RGB}{93,173,85}
\definecolor{myblue}{RGB}{30,90,100}
\definecolor{demphcolor}{RGB}{100,100,100}

\definecolor{datagreen}{RGB}{93,173,85}
\definecolor{datared}{RGB}{240,16,89}
\definecolor{datablue}{RGB}{0,114,188}

\newcolumntype{x}[1]{>{\centering\arraybackslash}p{#1pt}}
\newcolumntype{I}{!{\vrule width 1pt}}

\newcommand{\eg}{\textit{e.g.}\xspace}

\newcommand{\ie}{\textit{i.e.}\xspace}

\newcommand{\ourmethod}{\textsc{PromptCGL}\xspace}

\makeatletter
\newcommand{\thickhline}{%
	\noalign {\ifnum 0=`}\fi \hrule height 1pt
	\futurelet \reserved@a \@xhline
}

\makeatother

\makeatletter

\newcommand{\tablestyle}[4]{ 
	\centering
	\resizebox{#1\textwidth}{!}{
		\setlength\tabcolsep{#2pt}
		\renewcommand\arraystretch{#3}
		#4
	}
}
\makeatother

\begin{document}

\title{Prompt-Driven Continual Graph Learning}


\author{Qi Wang, Tianfei Zhou, Ye Yuan, and Rui Mao
\thanks{
	Qi Wang is with Beijing Institute of Technology, Zhuhai, China. 
	(e-mail: qiwang@bit.edu.cn)
	
	Tianfei Zhou and Ye Yuan are with the School of Computer Science and Technology, Beijing Institute of Technology, Beijing, China. 
	(e-mail: tfzhou@bit.edu.cn, yuan-ye@bit.edu.cn)
	
	Rui Mao is with the College of Computer Science and Software Engineering, Shenzhen University, Shenzhen, China. (e-mail: mao@szu.edu.cn)
}
}



\maketitle

\begin{abstract}
Continual Graph Learning (CGL), which aims to accommodate new tasks over evolving graph data without forgetting prior knowledge, is garnering significant research interest. Mainstream solutions adopt the memory replay-based idea, \ie, caching representative data from earlier tasks for retraining the graph model. However, this strategy struggles with scalability issues for constantly evolving graphs and raises concerns regarding data privacy. Inspired by recent advancements in the prompt-based learning paradigm, this paper introduces a novel prompt-driven continual graph learning (\ourmethod) framework, which learns a separate prompt for each incoming task and maintains the underlying graph neural network model fixed. In this way, \ourmethod naturally avoids catastrophic forgetting of knowledge from previous tasks. More specifically, we propose \textit{hierarchical prompting} to instruct the model from both feature- and topology-level to fully address the variability of task graphs in dynamic continual learning. Additionally, we develop a \textit{personalized prompt generator} to generate tailored prompts for each graph node while minimizing the number of prompts needed, leading to constant memory consumption regardless of the graph scale. Extensive experiments on four benchmarks show that \ourmethod achieves superior performance against existing CGL approaches while significantly reducing memory consumption. Our code is available at \url{https://github.com/QiWang98/PromptCGL}.
\end{abstract}


\section{Introduction}

Graphs are prevalent in numerous real-world applications, including social networks, biochemistry, and recommendation systems~\cite{yuan2011efficient, wu2020comprehensive, li2024cgmega, wang2022weighted}. Consequently, Graph Neural Networks (GNNs) have emerged as powerful tools for processing graph-structured data~\cite{wu2020comprehensive, xie2022semisupervised, wang2024noise, xu2024grakerformer}. However, traditional GNN methodologies typically assume static graph structures, which fail to capture the dynamic nature of the real world where graphs evolve continuously~\cite{liu2021overcoming, wang2022lifelong, zhang2022adaptive}. For instance, citation networks continually expand with the publication of new research papers, and co-purchasing networks grow as new categories of products are introduced. This necessitates models that can efficiently incorporate the features and topological information of new graphs in a continuous manner. 
\begin{figure}[t!]
	\begin{center}
		\subfloat[Conceptual illustration of replay-based method (left) and ours (right).]{\includegraphics[width=0.9\linewidth]{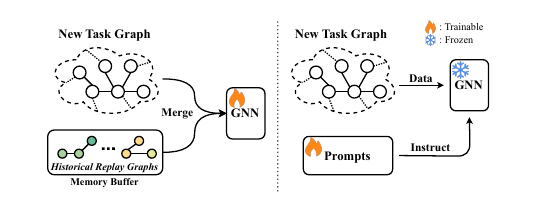}} \\
		\subfloat[Reddit Dataset] {\includegraphics[width=0.49\linewidth]{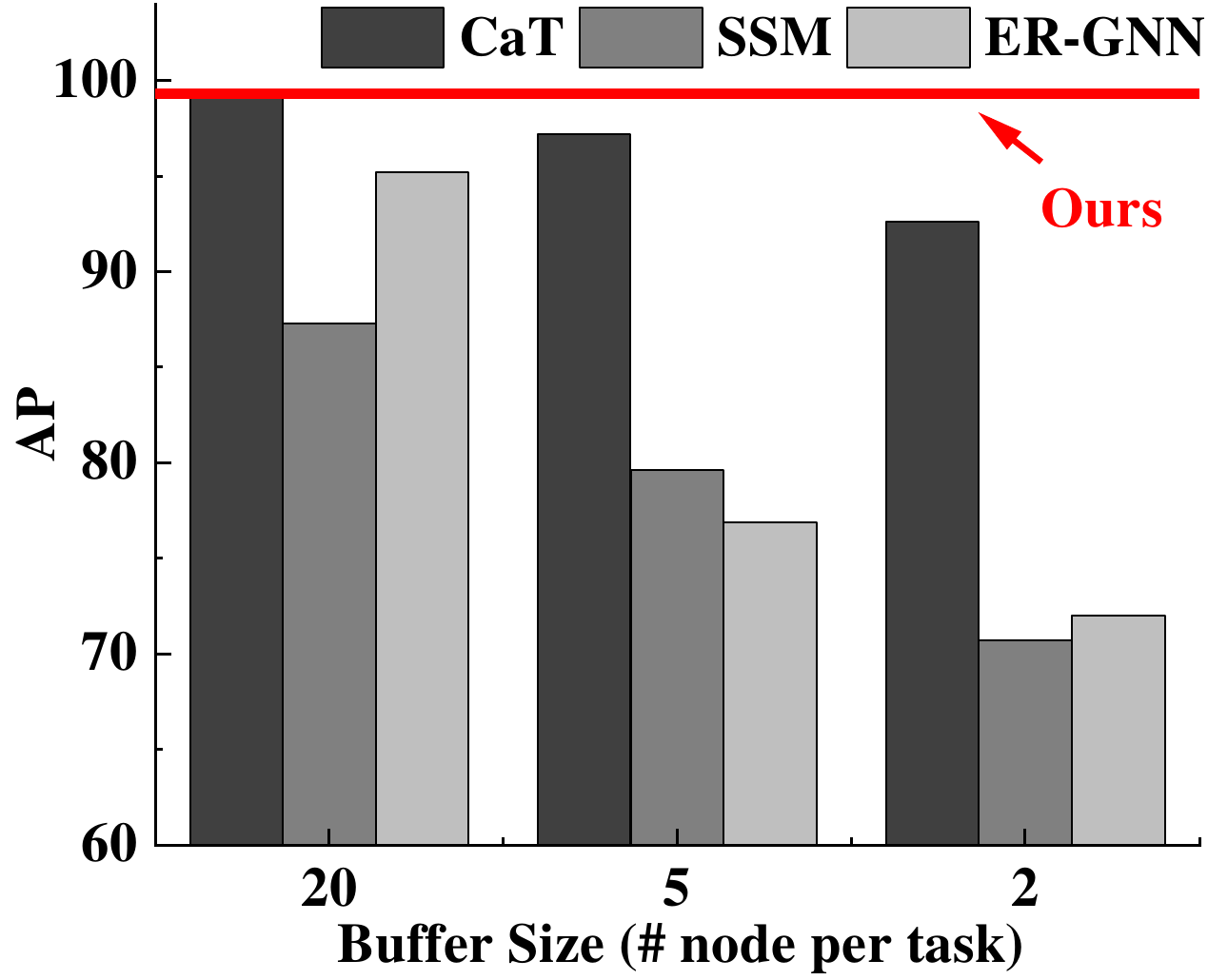}} \hfill
		\subfloat[Products Dataset] {\includegraphics[width=0.49\linewidth]{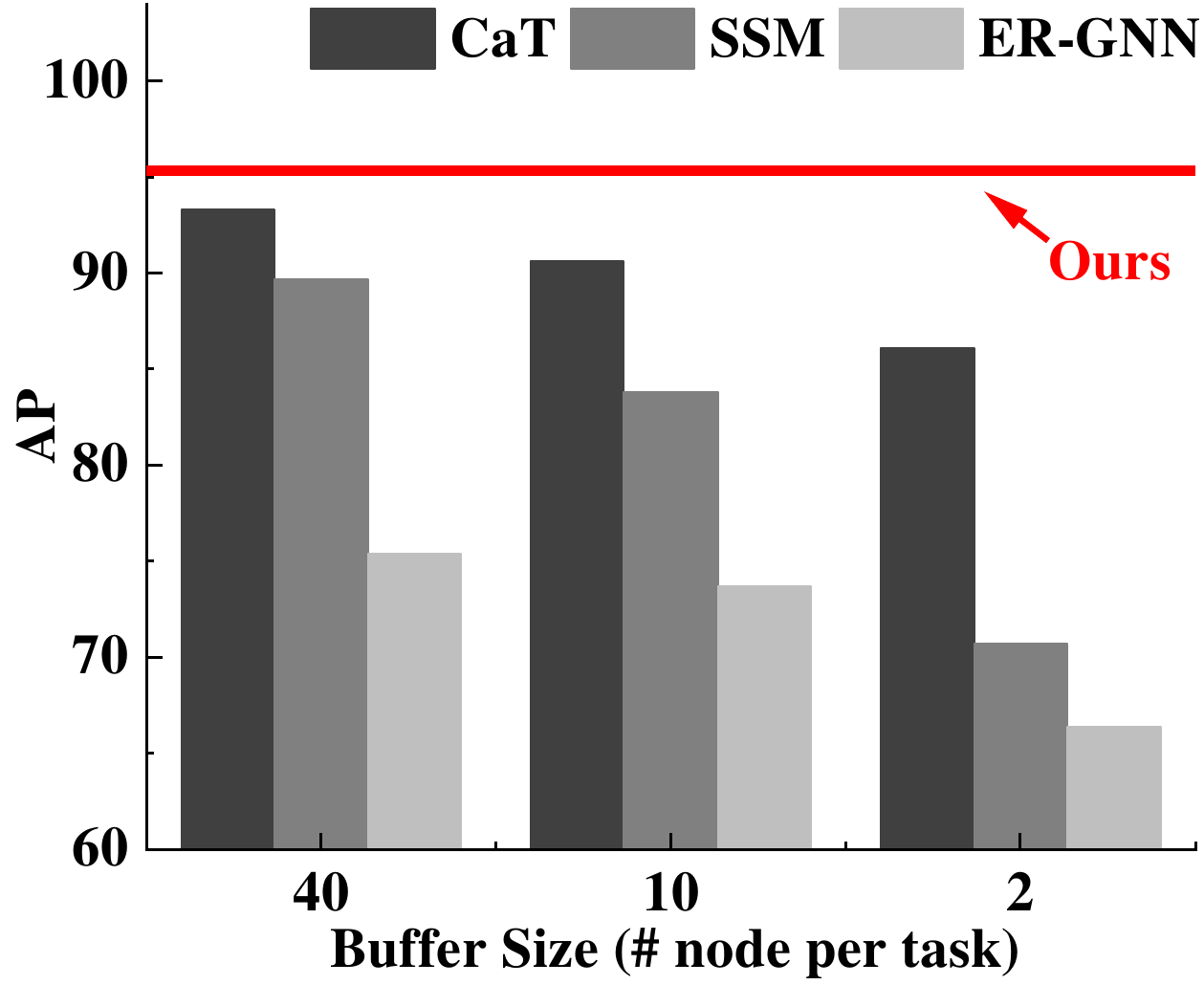}}
	\end{center}
	\caption{\textbf{Main Idea}. Replay-based methods, \eg, CaT~\cite{liu2023cat}, SSM~\cite{zhang2022sparsified}, ER-GNN~\cite{zhou2021overcoming}, require a memory buffer to store a number of graph nodes per task, which is merged  with the incoming graph for model retraining (see (a)). However, they face a severe degradation when the buffer size decreases (see (b) and (c)). In  contrast, \ourmethod represents a novel prompt-based learning paradigm, which learns a fixed number of prompts  for each unique task, and leaves GNNs parameters unchanged during the continual learning process. From (b) and (c), \ourmethod shows leading performance, regardless of the size of memory buffer.}
	\label{fig1}
\end{figure}

Due to the limitations in time overhead and computational resources, retraining the GNN models on entire datasets is impractical. Continual Graph Learning (CGL) thus emerges as a crucial paradigm to address the challenges posed by evolving graphs in the real world~\cite{zhang2022cglb, rakaraddi2022reinforced, cai2022multimodal, zhang2024continual}. Recent advancements in CGL can be broadly categorized into regularization~\cite{liu2021overcoming}, architectural design~\cite{zhang2022hierarchical, zhang2023continual}, and memory replay-based methods~\cite{zhou2021overcoming, zhang2022sparsified, kim2022dygrain}. Among these, replay-based methods have shown state-of-the-art  performance by storing sampled graphs in a memory buffer and replaying previous data while learning new tasks, as illustrated in Fig.~\ref{fig1} (a). Despite their effectiveness, replay-based CGL methods encounter two significant limitations. First, these methods demand substantial memory resources to store historical data, leading to performance degradation as buffer sizes decrease~\cite{liu2023cat, zhang2023ricci}, as demonstrated in Fig.~\ref{fig1} (b) and (c). Second, the storage of historical node information raises privacy concerns, especially in contexts involving sensitive data, such as purchase records in co-purchasing networks~\cite{wu2022federated}. These limitations indicate that simply buffering past data and retraining the model is not the optimal approach for CGL. 

With the success of foundation models, prompt learning has emerged as a key approach for transfer learning in large models~\cite{wang2022learning, xu2023making}. It shifts the focus from directly tuning model weights to designing prompts that effectively instruct the model to perform specific tasks while keeping the number of parameters fixed~\cite{li2021prefix, liu2023pre}. Recently, significant advancements in prompt learning for natural language processing and computer vision have inspired its application in graph learning~\cite{sun2023all, sun2023graph, fang2024universal, tan2023virtual, sun2022gppt}. These studies use graph prompts to bridge the gap between pre-trained pretext tasks and various downstream graph tasks on the same graph~\cite{tan2023virtual}. By leveraging easily obtainable graph information as pretext tasks, these methods pre-train GNNs, and then learn the prompts to reformulate other types of graph tasks as the pretext task~\cite{sun2022gppt}. This highlights the potential of prompt learning to transfer knowledge across different tasks on the graph.

However, the challenges posed by CGL are distinct, making existing graph prompt learning methods unsuitable for CGL scenarios. CGL requires the ability to continuously learn across \textbf{\textit{dynamic, incremental}} settings, where new task graphs introduce \textbf{\textit{unseen classes and varying topologies}}. This fundamentally differs from the \textbf{\textit{static, non-incremental}} setting of previous graph prompt learning methods, which focus on addressing task differences (\eg, node classification and edge prediction) within the same graph. Existing graph prompting methods are primarily designed to bridge the gap between different task types, without considering the variations in node features and topological structures that occur between different graphs. Additionally, these methods typically focus on transferring pre-trained models to downstream tasks, without addressing the critical issue of catastrophic forgetting, a central challenge in CGL setting.

Motivated by the above analysis, we propose a novel approach \ourmethod to explore prompt techniques in CGL. The basic idea of \ourmethod is to learn a set of unique prompts for each task to encode task-specific knowledge, and maintain graph neural networks mostly unchanged. This alleviates catastrophic forgetting and avoids privacy issues associated with retaining historical information. More concretely, we propose \textit{hierarchical prompting} to instruct the model from both feature- and topology-level to address the variability of task graphs in CGL fully. Additionally, to minimize memory consumption, \ourmethod develops a personalized prompt generator that produces personalized prompts based on different queries while maintaining a small prompt set, ensuring consistent memory usage regardless of graph scale. Furthermore, in the CGL setting, we implement a parameter-efficient prompt-tuning strategy: we freeze the pre-trained GNN weights of all layers except the prediction layer, updating only the prediction layer’s parameters and prompt-related parameters.

To summarize, our key contributions are as follows:
\begin{itemize}
	\item \ourmethod represents the first exploration of graph prompt learning in  CGL. It by nature mitigates catastrophic forgetting, reduces memory cost, and preserves data privacy. 
	
	\item We introduce a hierarchical prompting scheme to instruct the model to learn both node- and topology-aware specificity in the newly emerging tasks. 
	
	\item We develop a personalized prompt generator, which tailors prompts to individual nodes while maintaining minimal memory usage.
\end{itemize}

\section{Related Work}

\subsection{Continual Graph Learning}

Continual Graph Learning (CGL) aims to address the challenge of learning from a stream of graph-structured data over time while mitigating catastrophic forgetting. Current CGL methods can be broadly categorized into three main strategies: regularization methods~\cite{liu2021overcoming}, architectural design methods~\cite{zhang2022hierarchical, zhang2023continual}, and memory replay-based methods~\cite{zhou2021overcoming, kim2022dygrain, zhang2022sparsified, liu2023cat, zhang2023ricci}. Regularization methods focus on preventing catastrophic forgetting by adding constraints that help preserve knowledge across different learning tasks. Notable methods include TWP~\cite{liu2021overcoming}, which integrates topological information to retain learned features when adapting to new tasks. However, these methods often compromise the model's capacity to adapt efficiently to novel tasks, as the regularization can interfere with learning new knowledge. Architectural design approaches involve changes to the structure of the model to enhance its ability to learn and retain knowledge over time. For instance, HPNs~\cite{zhang2022hierarchical} introduce atomic feature extractors and hierarchical systems that scale dynamically to accommodate new knowledge. This approach increases model parameters and memory requirements as new tasks are added. Other architectural approaches focus on sparsity and modularization, which balance the trade-off between task retention and model expansion~\cite{zhang2023continual}. Memory replay-based methods are perhaps the most effective and widely studied, as they maintain a memory buffer to store data from previous tasks, which is replayed to mitigate forgetting while learning new tasks. A representative method is ER-GNN~\cite{zhou2021overcoming}, where experience replay involves storing sampled nodes and replaying them while learning new tasks. Other methods, such as SSM~\cite{zhang2022sparsified} and CaT~\cite{liu2023cat}, improve the efficiency of replay by using sparsified subgraphs or condensed graph modules to reduce replay memory overhead. Despite these advances, the scalability of replay-based methods is limited by increasing memory requirements, especially as the graph scale increases. 

In contrast to memory replay-based approaches, which typically store graph nodes at a ratio of 0.01, our approach introduces a novel method that significantly reduces memory overhead. By preserving only a small number of task-specific prompts (just 2 or 3), we enable the model to effectively learn the sequential tasks without incurring the memory overhead typical of traditional replay-based methods. Our approach achieves SOTA performance while minimizing memory costs, and addressing the scalability limitations of current CGL techniques.

\begin{figure*}[t]
	\centering
	\includegraphics[width=0.9\linewidth]{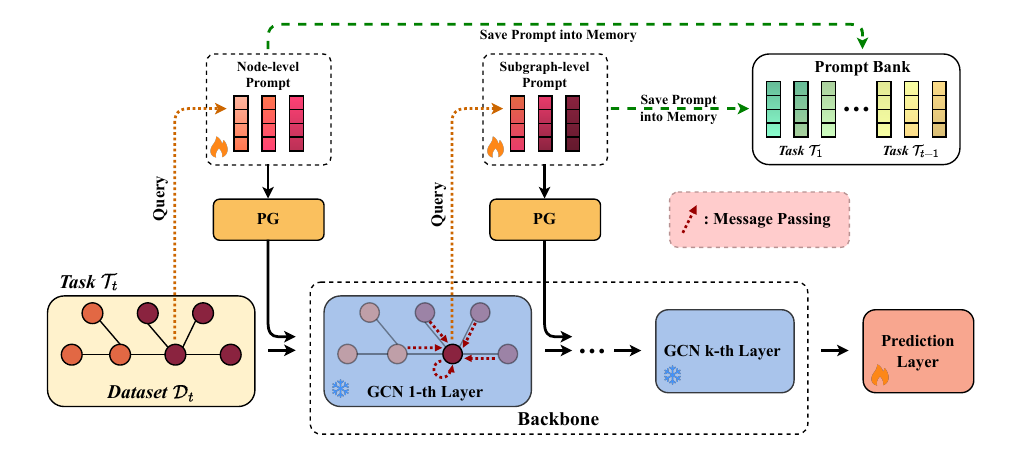}
	\caption{\textbf{Illustration of \ourmethod framework.} Here we present the execution steps for task $\mathcal{T}_t$. All tasks except $\mathcal{T}_0$ follow the same procedure. The backbone parameters, pre-trained on task $\mathcal{T}_0$, remain frozen in subsequent tasks. Initially, node-level personalized prompts are generated by the personalized prompt generator (PG) based on the query result of the node feature and a maintained small node-level prompt set, which are then added to the node features. These are processed through 1-th layer GNN to obtain node representations with topological information. Subsequently, subgraph-level personalized prompts are generated and added using the same method and passed into the subsequent networks. Learned prompts are saved into prompt bank after each task and selected based on task identity during inference for prediction.}
	\label{fig2}
\end{figure*}

\subsection{Prompt Learning}

Recently, prompt learning has emerged as a powerful technique in machine learning, particularly in natural language processing (NLP) and computer vision (CV). This approach has gained prominence due to its ability to adapt pre-trained large models to new tasks with minimal retraining, making it highly efficient for transfer learning scenarios~\cite{schick2021s, li2021prefix, zhou2022learning, zhou2022conditional, jia2022visual, chen2022adaptformer}. In NLP, prompt learning can be divided into two major categories: hard prompts and soft prompts. Hard prompts are manually crafted text additions, such as those used in PET-SGLUE~\cite{schick2021s}, where predefined templates are used to guide model predictions. Soft prompts, on the other hand, involve learnable vectors optimized for specific datasets, as seen in approaches like Prefix-tuning~\cite{li2021prefix}, which inserts task-specific vectors into the model while leaving the pre-trained parameters frozen. In CV, prompt learning often uses pre-defined visual prompts or learnable embeddings for task-specific adaptations~\cite{jia2022visual, chen2022adaptformer, deng2023prompt, hu2022promptcap, zheng2024exploring, du2022learning, li2024learning, luo2024vscode}, extending the versatility of visual transformer models to handle a wide variety of tasks, such as image captioning~\cite{deng2023prompt, hu2022promptcap}, classification~\cite{zheng2024exploring}, and object detection~\cite{du2022learning, li2024learning, luo2024vscode}.

Despite its success in both NLP and CV, the unique characteristics of graph data—such as its non-sequential nature and complex relational structure—present challenges for directly applying these prompt learning techniques. As a result, there has been a growing interest in developing specialized prompt learning techniques tailored to graph data.

\subsection{Graph Prompt Learning}

Existing research in graph prompt learning has predominantly focused on customizing pre-training tasks and leveraging designed prompts to address various graph tasks within static graphs, such as node classification, edge prediction, and graph classification~\cite{sun2022gppt, tan2023virtual, sun2023all, sun2023graph, liu2023graphprompt, fang2024universal}. The goal of these methods is to reformulate downstream tasks on a static graph in a manner that aligns with pretext tasks, allowing models to generalize better across multiple tasks. GPPT~\cite{sun2022gppt} was one of the pioneering works in this area, enhancing GNN generalization capabilities through a combination of graph pre-training and prompt-tuning. The GPPT method begins with pre-training on link prediction tasks and then reformulates node classification tasks as link predictions between target nodes and category nodes, leveraging prompts to adapt to the specific task at hand. This approach allows the model to learn a shared representation of nodes and edges that is transferable across different tasks. Building on this idea, GPF~\cite{fang2024universal} introduces a universal prompt-tuning method, which fine-tunes only a small subset of parameters across various graph tasks, allowing for task-specific adaptations without retraining the entire model. The ``All in One" approach~\cite{sun2023all} further extends this by reformulating node and edge prediction tasks as subgraph-level tasks and designing multi-task prompts using meta-learning techniques to handle a variety of graph tasks. These methods have made significant strides in bridging the gap between different tasks within the same graph. However, most of these approaches focus on static graphs, where the topology and feature distribution remain fixed.

While static graph-based methods excel in scenarios where the graph structure and features do not change, they fall short when applied to dynamic, incremental graphs. The topology and feature distributions in such graphs are continually evolving, posing a unique challenge in maintaining performance across tasks that span different graphs or across time. In contrast to these static graph-based approaches, we focus on extending graph prompt learning to dynamic graph settings. Our method addresses the feature and topology gaps between different task graphs in dynamic, evolving environments, providing a more flexible and scalable solution for CGL scenarios.

\section{Methodology}

\subsection{Problem Definition}

A typical setup for CGL involves training the model on multiple tasks with non-overlapping classes that arrive sequentially. Denote  $n$ tasks as $\mathcal{T}=\{\mathcal{T}_0,\mathcal{T}_1,\ldots,\mathcal{T}_n\}$ and corresponding sequence of datasets as $\mathcal{D}=\{\mathcal{D}_1, \mathcal{D}_2, \ldots, \mathcal{D}_n\}$. In this continual learning paradigm, the model only has access to the dataset $\mathcal{D}_t$ of the current task $\mathcal{T}_t$, while datasets from prior tasks ($\mathcal{D}_i \mid i < t$) are unavailable. In node classification scenarios, each dataset $\mathcal{D}_t$ consists of a graph $\mathcal{G}_t = (\mathcal{V}_t, \mathcal{E}_t)$, where $\mathcal{V}_t$ represents the set of nodes and $\mathcal{E}_t$ denotes the set of edges in the graph. The number of nodes in the graph is $N = |\mathcal{V}_t|$, and each node is assigned a label from a set of node labels $\mathcal{Y}_t$. An ideal CGL method should achieve optimal performance on the current task while maintaining the performance on previous tasks.

\subsection{Prompt Driven Continual Graph Learning}

\subsubsection{Overview} Fig.~\ref{fig2} shows the framework of our \ourmethod. Our model consists of three main parts: a backbone $g_{\Theta}$, consisting of a multilayer GNN for feature extraction, a prediction layer $f_{\Phi}$ for performing classification tasks, and the prompts $\mathbf{P}$. During training, we first pre-train the backbone $g_{\Theta}$ and the prediction layer $f_{\Phi}$ on the initial task $\mathcal{T}_0$ without prompts and then freeze the backbone parameters in subsequent tasks to ensure model stability and consistency. The frozen backbone serves as a feature extractor for obtaining node representations with topological information in later tasks. The core of  \ourmethod is to find  the optimal prompts $\mathbf{P}$ for each task by solving the following objective: 
\begin{equation}
	\max_{\Phi,\mathbf{P}} \mathbb{E}_{(\mathbf{X}_0, \mathbf{A}, y) \sim \mathcal{T}_i} \left[P(y | f_{\Phi}(g_{\Theta}(\mathbf{X}_0, \mathbf{A}, \mathbf{P}))) \right],
	\label{eq1}
\end{equation}
where $\mathbf{X}_0 \in \mathbb{R}^{N\times d_f}$ indicates the node features, $\mathbf{A}\in \mathbb{R}^{N\times N}$ is the adjacency matrix, and $\mathbf{P}$ consists of node-level prompts $\mathbf{P}_n$ and subgraph-level prompts $\mathbf{P}_s$, both in $\mathbb{R}^{k \times d_f}$ and composed of $k$ independent prompt vectors, where $d_f$ denotes the feature dimension. In the continual learning process, \textit{\textbf{only the parameters of $f_{\Phi}$ and prompts $\mathbf{P}$ are learnable}}. Upon completion of each task, we save the prompts into a prompt bank, which would be retrieved at inference time. Next, we present two core techniques of \ourmethod, including hierarchical prompting (HP) and a personalized prompt generator (PG).

\subsubsection{Hierarchical Prompting}

In a continual learning setting, task graphs consist of nodes with non-overlapping classes, leading to significant differences in both features and topology structures between the target and initial task graphs. For example, in the case of social networks representing distinct interest groups, one graph may correspond to a community centered on entertainment, while another could represent a group focused on computer science. These communities differ not only in their feature distributions—shaped by the distinct interests and behaviors of their members—but also in their topologies. An entertainment-based community might form tightly-knit clusters, while a technology-focused group could exhibit a more dispersed and expansive network. Such disparities in both feature space and network structure complicate knowledge transfer across tasks. To address these challenges, we propose a hierarchical prompting strategy, which involves node-level prompts to address feature discrepancies and subgraph-level prompts to handle structural variations.

For node-level prompts, we generate personalized prompts for each node based on the initial node features $\mathbf{X}_0$ and the maintained node-level prompts $\mathbf{P}_n$ using the PG component. These personalized prompts are then added to the node features to obtain the prompted node features $\mathbf{X}_0^p$.
\begin{equation}
	\mathbf{X}_0^p = \mathbf{X}_0 + \textit{PG} (\mathbf{X}_0, \mathbf{P}_n),
	\label{eq2}
\end{equation}
where $\textit{PG}$ denotes the personalized prompt generator, whose output shares the same dimensions as node features $\mathbf{X}_0$.

For subgraph-level prompts, we first encode the prompted node features and the relationships between neighboring nodes using the frozen first-layer GNN, resulting in node representations that incorporate topological information $\mathbf{X}_1$:
\begin{equation}
	\mathbf{X}_1 = \textit{GNN}_1 (\mathbf{X}_0^p, \mathbf{A}),
	\label{eq3}
\end{equation}
where $\textit{GNN}_1$ is the first-layer GNN of the backbone, $\mathbf{X}_1 \in \mathbb{R}^{N \times d_h}$, and $d_h$ is the hidden dimension. Then following a similar procedure to the node-level prompts, we generate the prompted node representations $\mathbf{X}_1^p \in \mathbb{R}^{k \times d_h}$ by utilizing subgraph-level prompts $\mathbf{P}_s$ and the node representations with topological information $\mathbf{X}_1$:
\begin{equation}
	\mathbf{X}_1^p = \mathbf{X}_1 + \textit{PG} (\mathbf{X}_1, \mathbf{P}_s),
	\label{eq4}
\end{equation}
The resulting prompted node representations, $\mathbf{X}_1^p$, are then passed on to the subsequent layers of the network for further processing.

\begin{figure}[t]
	\centering
	\includegraphics[width=0.7\linewidth]{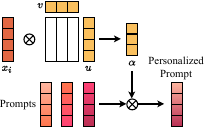}
	\caption{Illustration of the Personalized Prompt Generator.}
	\label{fig_PG}
\end{figure}

\subsubsection{Personalized Prompt Generator}\label{PG}

The inherent heterogeneity of nodes implies that using a uniform prompt for all nodes leads to inefficiency and poor performance. Although customizing unique prompts for each node is ideal, this approach would significantly increase memory consumption, particularly as the graph scales. To mitigate this, we propose a personalized prompt generator, as illustrated in Fig.~\ref{fig_PG}. This generator utilizes the unique representation of each node and a query matrix to derive personalized prompts.

Specifically, for node $i$, the personalized prompt $\mathbf{p}^p_i$ is generated by dynamically aggregating the maintained prompts based on the query result of its representation $\boldsymbol{x}_i$:

\begin{equation}
	\mathbf{p}^p_i = \sum\nolimits_{j=0}^k {\alpha}_j \cdot \boldsymbol{p}_j, \quad 
	\boldsymbol{\alpha} = \text{Softmax} (\mathbf{Q} \boldsymbol{x}^0_i),
	\label{eq5}
\end{equation}
where $\mathbf{Q}$ is a query matrix used to compute importance weights $\alpha$ of each prompt to the node $i$. These weights, which differ across nodes, enable the creation of personalized prompts through tailored aggregation.

To enhance model stability and improve computational and storage efficiency, we decompose the query matrix into two specific low-order vectors: node-dependent $\mathbf{v}\in \mathbb{R}^{1\times k}$ and prompt-dependent $\mathbf{u}\in \mathbb{R}^{1\times n}$.
\begin{equation}
	\mathbf{Q} = \boldsymbol{u} \otimes \boldsymbol{v}.
	\label{eq6}
\end{equation}

\subsubsection{Learning Objective}
For the current tasks, both the prompts and node features $\mathbf{X}_0$ are input into the backbone network $g$, followed by a predictive layer to generate the final outputs. The model is trained end-to-end using the following loss function:
\begin{equation}
	\min_{\Phi, \mathbf{P}_n, \mathbf{P}_s} \mathcal{L} \Big ( f_{\Phi} \big(g_{\Theta} \left( \mathbf{X}_0, \mathbf{A}, \mathbf{P}_n, \mathbf{P}_s \right) \big) \Big ), \quad \mathbf{X} \in \mathcal{D}_t,
	\label{eq7}
\end{equation}
where $\mathcal{L}$ denotes the cross-entropy loss. 

To further mitigate the forgetting ratio, a smaller learning rate is applied to the predictive layer compared to the learning rates used for the HP and PG components during training. This adjustment is necessary because, in CGL, the predictive layer is shared across all tasks. A lower learning rate ensures the stability and generalization of this layer over multiple tasks. Upon completion of the current task, the parameters associated with the learned prompts are stored in the prompt bank for future inference. The training process of \ourmethod is comprehensively outlined in Algorithm~\ref{alg}, providing a detailed overview of the method.

\subsubsection{Inference}
For the $i$-th inference task, the HP and PG are initialized by retrieving the corresponding prompt parameters for the $i$-th task from the prompt bank based on the task identifier. Inference is then conducted to obtain the predicted result $\hat{y}$ as follows: 
\begin{equation}
	\hat{y} = f_{\Phi}\left (g_{\Theta} \left (\mathbf{X}_0, \mathbf{A}, \mathbf{P}^i \right ) \right),
	\label{eq8}
\end{equation}
where $\mathbf{P}^i = (\mathbf{P}_s, \mathbf{P}_n)$ represents the prompt parameters associated with the $i$-th task. 

\begin{algorithm}[t]
	\renewcommand{\algorithmicrequire}{\textbf{Input:}}
	\renewcommand{\algorithmicensure}{\textbf{Output:}}
	\begin{algorithmic}[1]
		\REQUIRE Dataset \(\mathcal{D}\), a pre-trained backbone \(g_{\Theta}\), and a prediction layer \(f_{\Phi}\).
		\ENSURE Prompt Bank \(\mathcal{P}\).
		\STATE Initialize the memory bank \(\mathcal{P} \leftarrow \emptyset\);
		\FOR{each task \(\mathcal{D}_i\) in \(\mathcal{D}\)}
		\STATE Randomly initialize node-level prompts \(\mathbf{P}_{i,n}\) and subgraph-level prompts \(\mathbf{P}_{i,s}\) for task \(i\);
		\STATE Compute node-level personalized prompts \(\mathbf{P}_{i,n}\) using Eq.~(\ref{eq5}) and Eq.~(\ref{eq6}) with input \(\mathbf{X}_0 \in \mathcal{D}_i\);
		\STATE Add the node-level personalized prompts \(\mathbf{P}_{i,n}\) to \(\mathbf{X}_0\) to obtain the prompted node features \(\mathbf{X}_0^p\) (Eq.~(\ref{eq2}));
		\STATE Compute node representations with topological information \(\mathbf{X}_1\) using Eq.~(\ref{eq3}) with the prompted node features \(\mathbf{X}_0^p\);
		\STATE Compute subgraph-level personalized prompts \(\mathbf{P}_{i,s}\) using Eq.~(\ref{eq5}) and Eq.~(\ref{eq6}) with the node representations \(\mathbf{X}_1\);
		\STATE Add the subgraph-level personalized prompts \(\mathbf{P}_{i,s}\) to \(\mathbf{X}_1\) to obtain the prompted node representations \(\mathbf{X}_1^p\) (Eq.~(\ref{eq4}));
		\STATE Update prompts and prediction layer using Eq.~(\ref{eq7});
		\STATE Store the learned prompts \(\mathbf{P}_{i,n}\) and \(\mathbf{P}_{i,s}\) into the memory bank \(\mathcal{P}\);
		\ENDFOR
		\RETURN Prompt Bank \(\mathcal{P}\);
	\end{algorithmic}
	\caption{The Process of Training \ourmethod}
	\label{alg}
\end{algorithm}

\subsection{Algorithm}

In order to present our method more systematically, the process of training PromptCGL is outlined in Algorithm~\ref{alg}. 

The algorithm takes as input the dataset \(\mathcal{D}\), a pre-trained backbone \(g_\Theta\), and a prediction layer \(f_\Phi\). It initializes a memory bank \(\mathcal{P}\) to store the learned prompts (Line~2). For each task \(\mathcal{D}_i\) in the dataset, node-level prompts \(\mathbf{P}_{i,n}\) and subgraph-level prompts \(\mathbf{P}_{i,s}\) are randomly initialized (Line~4). The node-level prompts are then personalized for each node in the graph using Eq.~\eqref{eq5} and Eq.~\eqref{eq6} (Line~5), and these prompts are added to the node features to generate the prompted node features \(\mathbf{X}_0^p\) (Line~6). These features are passed through the backbone's first layer to compute topologically enriched node representations \(\mathbf{X}_1\) (Line~7). Next, subgraph-level prompts \(\mathbf{P}_{i,s}\) are personalized using the same mechanism as the node-level prompts (Line~8), and the updated node representations are computed (Line~9). The prompts and the prediction layer parameters are updated using the loss function defined in Eq.~\eqref{eq7} (Line~10). Finally, the learned prompts \(\mathbf{P}_{i,n}\) and \(\mathbf{P}_{i,s}\) are stored in the memory bank for future retrieval (Line~11). This process enables the model to adapt to new tasks while preserving knowledge from previous tasks, ensuring efficient and scalable continual graph learning. At the end of the process, the memory bank \(\mathcal{P}\), containing all task-specific prompts, is returned (Line~12).

\subsection{Discussion}


\ourmethod requires only a small number of prompts and low-order query vectors, and these quantities remain constant regardless of the graph scale. Specifically, the space complexity for maintaining two levels of prompts is $O(k(d_f+d_h))$, where $d_f$ and $d_h$ are the dimensions of the feature and hidden layers, respectively. The space complexity for low-order query vectors is $O(d_f+d_h)$, leading to a total space complexity of $O(k \cdot d)$. Experiments in TABLE~\ref{main_result} confirm that the model achieves SOTA performance with $k$ set to just 3. In contrast, replay-based methods require sampling based on a ratio $k_{rate}$, resulting in a space complexity of $O(k_{rate} \cdot N \cdot d)$, \ie, $O(N \cdot d)$. This linear relationship with graph scale means replay-based methods demand significantly more memory for large-scale graphs, whereas our method maintains minimal memory consumption.

Furthermore, \ourmethod demonstrates lower training costs compared to regularization and replay-based methods. Unlike regularization techniques, which require the computation of an additional loss function, and replay-based methods, which necessitate retraining all parameters and the inclusion of extra sampling modules, our approach focuses solely on fine-tuning the parameters associated with prompts and prediction layers. This streamlined process results in significantly reduced computational overhead. 

Additionally, Our method offers superior privacy-preserving capabilities over replay-based methods by storing prompts instead of historical data, providing a more secure and reliable solution for integrating with GNNs in sensitive scenarios.


\section{Experiments}

\subsection{Datasets}

Following~\cite{zhang2022cglb}, we conduct extensive experiments on four  public datasets: CoraFull~\cite{mccallum2000automating}, OGB-Arxiv~\cite{hu2020open}, Reddit~\cite{hamilton2017inductive}, and OGB-Products~\cite{hu2020open}. TABLE~\ref{dataset_table} shows their detailed statistics. For all datasets, the data of each class is divided into 60\% for training, 20\% for validation, and the remaining 20\% for testing, and each task includes data from two classes. 

The detailed descriptions of these datasets are as follows:

\begin{enumerate}
	\item CoraFull~\cite{mccallum2000automating} is an extension of the well-known Cora dataset. Nodes represent scientific publications and edges represent citation links between them.
	\item OGB-Arxiv~\cite{hu2020open} is part of the Open Graph Benchmark (OGB). Nodes represent Computer Science arXiv papers, and edges denote citation relationships. Each node is associated with a 128-dimensional feature vector derived from the paper's title and abstract.
	\item Reddit~\cite{hamilton2017inductive} is a graph dataset where nodes correspond to posts in the Reddit social network, and edges represent interactions between these posts.
	\item OGB-Products~\cite{hu2020open} is another dataset from the OGB. Nodes represent products sold on Amazon, and edges indicate that two products are frequently bought together.
\end{enumerate}

\begin{table}[t]
	\centering
	\caption{Detailed statistics of four datasets.}
	\hspace{-1.5em}
	\tablestyle{0.5}{3}{1.1}{
		\begin{tabular}{c||c|c|c|c}
			\thickhline
			\rowcolor{mygray}
			\textbf{Dataset} & \textbf{CoraFull} & \textbf{OGB-Arxiv} & \textbf{Reddit} & \textbf{OGB-Products} \\
			\hline \hline
			\textbf{Nodes}    & 19,793   & 169,343   & 227,853     & 2,449,028    \\
			\textbf{Edges}    & 130,622  & 1,166,243 & 114,615,892 & 61,859,036   \\
			\textbf{Features} & 8,710    & 128       & 602         & 100          \\
			\textbf{Classes}  & 70       & 40        & 40          & 46           \\
			\textbf{Tasks}    & 35       & 20        & 20          & 23           \\
			\hline
		\end{tabular}
	}
	\label{dataset_table}
\end{table}

\subsection{Baselines}

For a comprehensive assessment, \ourmethod is contrasted with state-of-the-art  baselines. These encompass four traditional continual learning methods: \textit{EWC}~\cite{kirkpatrick2017overcoming}, \textit{MAS}~\cite{aljundi2018memory}, \textit{GEM}~\cite{lopez2017gradient}, and \textit{LwF}~\cite{li2017learning}, along with five CGL methods: \textit{TWP}~\cite{liu2021overcoming}, \textit{HPNs}~\cite{zhang2022hierarchical}, \textit{ER-GNN}~\cite{zhou2021overcoming}, \textit{SSM}~\cite{zhang2022sparsified}, and \textit{CaT}~\cite{liu2023cat}. Additionally, two baselines are established: \textit{Bare}, a lower-bound baseline that is directly fine-tuned on the task sequence without CGL techniques, and \textit{Joint}, an ideal upper-bound baseline that utilizes data from all historical tasks during new task learning. The details of the baseline continual learning methods are as follows:

\begin{enumerate}
	\item \textbf{Bare} is fine-tuned directly on the task sequence without any CGL methods to avoid forgetting, therefore we regard it as a lower bound on continual graph learning performance.
	\item \textbf{EWC}~\cite{kirkpatrick2017overcoming} protects important weights for previously learned tasks by penalizing their updates during new task learning.
	\item \textbf{MAS}~\cite{aljundi2018memory} introduces a memory module for slowing down the updating of important parameters by measuring their importance based on sensitivity to their prediction.
	\item \textbf{GEM}~\cite{lopez2017gradient} stores representative data in episodic memory and modifies the gradients using the informative data stored in memory.
	\item \textbf{LwF}~\cite{li2017learning} preserves the knowledge of the model in the new model through knowledge distillation.
	\item \textbf{TWP}~\cite{liu2021overcoming} preserves the topological information of previous tasks by regularisation penalties.
	\item \textbf{HPNs}~\cite{zhang2022hierarchical} use atomic feature extractors and a hierarchical prototype system for continual learning.
	\item \textbf{ER-GNN}~\cite{zhou2021overcoming} integrates memory-replay to GNNs by sampling the informative nodes from the previous task graph into the memory buffer.
	\item \textbf{SSM}~\cite{zhang2022sparsified} stores the sparsified previous task graph in the memory buffer for replay.
	\item \textbf{CaT}~\cite{liu2023cat} stores the condensed previous task graph in the memory buffer and utilizes a train in the memory scheme to update the model.
	\item \textbf{Joint} utilizes entire data  of all historical tasks to update the model when learning new tasks, therefore we regard it as an upper bound on continual graph learning performance.
\end{enumerate}

\subsection{Evaluation Metrics}
Following the methodology in~\cite{zhang2022cglb} and~\cite{liu2023cat}, we evaluate model performance using two metrics: \textit{average performance} (AP) and \textit{average forgetting} (AF). Given a sequence of $T$ tasks, the accuracy of the model on the $q$-th task after learning the $p$-th task is denoted as $m_{p,q}$. The set ${m_{T,q} | q=0, 1, \dots, T-1}$ represents the accuracy of each learned task after completing the entire sequence of tasks. Formally, the definitions for AP and AF are as follows:
\begin{equation}
	\mathrm{AP} = \frac{\sum_{q = 1}^{T} m_{T, q}}{T},  \quad \mathrm{AF} = \frac{\sum_{q = 1}^{T-1}\left(m_{T, q}-m_{q, q}\right)}{T-1}.
\end{equation}
We utilize AP to assess the overall performance across all tasks at the end of the task sequence. AP is considered a more critical metric compared to AF, which quantifies the extent of forgetting during the continual learning process.

\subsection{Implementation} 

Following~\cite{zhang2022sparsified} and~\cite{liu2023cat}, we employ a two-layer GCN~\cite{kipf2016semi} as the backbone for all models, except for TWP, which uses GAT~\cite{velivckovic2017graph} due to its attention-based mechanism for assessing topological importance. The hidden layer dimension for all GNNs is set to 32. During the pre-training phase, we use the Adam optimizer with a learning rate of 0.001 and a weight decay of 5e-4. In the continual learning phase, we apply different learning rates: 5e-4 without weight decay for the prediction layer, and 0.01 with a weight decay of 5e-4 for the prompts. 

All experimental results are reported as the mean and standard deviation across three independent runs. The experiments were conducted on a machine equipped with two Tesla T4 GPUs, each with 16 GB of memory, and an Intel(R) Xeon(R) Silver 4210 CPU @ 2.20 GHz.

\begin{table*}[t!]
	\centering
	\caption{Performance comparisons between \ourmethod and baselines on four datasets. The best and second-best results are shown in bold and underlined, respectively. $^*$: results of HPNs are from the original paper.}
	\resizebox{\textwidth}{!}{
		\setlength\tabcolsep{6pt}
		\renewcommand\arraystretch{1.4}
		\begin{tabular}{c||c|cc|cc|cc|cc}
			\thickhline
			\rowcolor{mygray}
			 &  & \multicolumn{2}{c|}{\textbf{CoraFull}} & \multicolumn{2}{c|}{\textbf{Arxiv}} & \multicolumn{2}{c|}{\textbf{Reddit}} & \multicolumn{2}{c}{\textbf{Products}} \\
			\cline{3-10}
			\rowcolor{mygray}
			\multicolumn{1}{c||}{\multirow{-2}{*}{\textbf{Category}}} & \multicolumn{1}{c|}{\multirow{-2}{*}{\textbf{Methods}}} & AP(\%) & AF(\%) & AP(\%) & AF(\%) & AP(\%) & AF(\%) & AP(\%) & AF(\%) \\
			\hline \hline
			{Lower bound} & {Bare} & 61.4±2.3 & -33.1±3.5 & 74.4±1.6 & -6.8±3.4 & 65.5±5.2 & -35.8±5.4 & 68.6±1.9 & -28.1±2.3 \\
			\hline
			& EWC & 88.6±2.0 & -7.2±2.0 & 89.7±3.2 & -6.1±3.5 & 76.4±8.2 & -24.4±8.7 & \underline{91.2±0.2} & -3.5±0.2 \\
			& MAS & 74.9±1.7 & \underline{-0.4±0.1} & \underline{90.1±0.7} & -6.0±0.9 & \underline{98.9±0.1} & -0.1±0.0 & 89.5±0.9 & \underline{-0.1±0.3} \\
			& GEM & 86.4±0.3 & -9.4±0.7 & 81.6±5.1 & -5.6±6.0 & 94.2±2.0 & -5.7±2.1 & 84.8±0.4 & -11.8±0.3 \\
			\multicolumn{1}{c||}{\multirow{-4}{*}{{Regularization}}} & TWP & 86.9±1.5 & -4.4±0.7 & 73.2±4.7 & -1.2±0.6 & 74.1±5.5 & -1.5±0.5 & 75.5±4.4 & -4.9±6.4 \\
			\hline
			{Distillation} & LwF & 62.1±2.7 & -27.6±5.4 & 74.0±0.4 & -9.7±4.3 & 65.1±4.7 & -36.3±5.0 & 61.6±1.2 & -35.5±1.3 \\
			\hline
			{Architecture} & HPNs$^*$ & - & - & 85.8±0.7 & \textbf{0.6±0.9} & - & - & 80.1±0.8 & \textbf{2.9±1.0} \\
			\hline
			\rowcolor{lightgray}
			\multicolumn{10}{c}{\textbf{setting: 2 prompts in \ourmethod} ($=3.0/3.8/3.2/4.0$ nodes in CoraFull/Arxiv/Reddit/Products)} \\
			\hline
			& ER-GNN & 64.9±1.7 & -31.5±0.9 & 74.7±2.4 & -13.3±2.9 & 67.9±5.2 & -33.3±5.4 & 70.7±3.7 & -26.7±3.8 \\
			& SSM & 77.9±1.2 & -18.6±0.9 & 77.8±2.7 & -10.6±2.4 & 75.9±3.6 & -25.0±3.8 & 77.1±1.1 & -19.8±1.4 \\
			\multicolumn{1}{c||}{\multirow{-3}{*}{{Replay}}} & CaT & 93.1±0.4 & -2.9±0.8 & 73.7±2.8 & -12.6±4.7 & 90.6±1.4 & -9.4±1.5 & 88.5±0.4 & -8.2±0.5 \\
			\hline
			{Prompt} & \textbf{Ours} & \underline{94.8±0.7} & -0.9±0.6 & 96.3±0.6 & -0.5±0.5 & 99.5±0.0 & \textbf{-0.0±0.0} & 94.9±0.6 & -0.6±0.3 \\
			\hline
			\rowcolor{lightgray}
			\multicolumn{10}{c}{\textbf{setting: 3 prompts in \ourmethod} ($=4.0/5.0/4.2/5.3$ nodes in CoraFull/Arxiv/Reddit/Products)} \\
			\hline
			& ER-GNN & 65.2±4.9 & -31.3±4.6 & 75.6±3.6 & -12.5±2.0 & 69.2±5.4 & -32.0±5.7 & 69.7±1.3 & -27.8±1.6 \\
			& SSM & 80.2±0.9 & -16.1±0.8 & 74.7±2.4 & -13.3±2.9 & 87.2±2.7 & -13.1±2.8 & 79.6±1.4 & -17.3±1.5 \\
			\multicolumn{1}{c||}{\multirow{-3}{*}{{Replay}}} & CaT & 93.5±0.3 & -2.7±0.7 & 75.7±2.5 & -12.4±0.6 & 93.5±1.6 & -6.5±1.7 & 88.4±0.3 & -8.2±0.4 \\
			\hline
			{Prompt} & \textbf{Ours} & \textbf{95.4±0.6} & \textbf{-0.3±0.2} & \textbf{96.7±0.1} & \underline{-0.1±0.1} & \textbf{99.3±0.1} & \underline{-0.2±0.1} & \textbf{95.3±0.2} & -0.2±0.4 \\
			\hline
			{Upper bound} & \textit{Joint} & 96.1±0.1 & 0.5±0.4 & 95.6±0.4 & 0.0±0.3 & 99.6±0.0 &  0.0±0.0 & 95.5±0.2 & -0.9±0.1 \\
			\hline
		\end{tabular}
	}
	\label{main_result}
\end{table*}

\subsection{Analysis of Main Results}

\subsubsection{Performance Comparison}

Table~\ref{main_result} presents a comparative analysis across four benchmark datasets. For the replay-based methods, we conducted experiments under two distinct scenarios, ensuring that memory consumption was aligned with \ourmethod for a fair comparison. In particular, the setting in TABLE~\ref{main_result} shows under what circumstances the memory consumption of storing prompts in our method equals that of storing nodes in the replay-based methods.

The results demonstrate that \ourmethod consistently achieves SOTA performance across all benchmarks with minimal memory consumption. Notably, \ourmethod reaches the ideal upper bound of joint training performance on all datasets with only three prompts. On the Arxiv dataset, \ourmethod even surpasses the joint training method, showing that the prompts in our approach can push model performance beyond traditional upper bounds. Although the AF for \ourmethod on the Arxiv and Products datasets are not optimal—0.1\% and 0.2\%, respectively—these minor rates have negligible impact on the overall model performance. Furthermore, \ourmethod demonstrates robust performance with just two prompts, achieving the highest average accuracy across all datasets while keeping average forgetting ratios consistently below 1\%. This highlights \ourmethod's ability to adapt to new tasks while effectively preserving historical knowledge, thus mitigating catastrophic forgetting.

In contrast, other baseline methods are unable to match the performance of \ourmethod. Regularization techniques such as EWC, MAS, GEM, and TWP, along with distillation-based approaches like LwF, impose additional constraints on the model while learning new tasks to mitigate forgetting. However, these constraints often limit the model's plasticity, resulting in suboptimal performance on new tasks despite some methods achieving low forgetting ratios. Replay-based methods, as demonstrated in previous studies~\cite{zhang2022sparsified, liu2023cat}, can approach joint training performance with a storage ratio of $k_{rate} = 0.01$. Nonetheless, this level of memory consumption is prohibitively high for large-scale graphs. When replay-based methods are compelled to minimize memory usage by retaining only minimal historical information, they experience catastrophic forgetting, as shown in Table~\ref{main_result}. Under these constrained conditions, their performance deteriorates below that of regularization methods. Consequently, replay-based approaches are unsuitable for large-scale graphs, where memory constraints are more stringent. In contrast, \ourmethod sustains robust performance with a low memory footprint, effectively mitigating catastrophic forgetting without the necessity for extensive memory buffers, thereby demonstrating greater scalability for large-scale graph tasks.

\begin{figure*}[t!]
	\centering
	\subfloat[CoraFull Dataset]{\includegraphics[width=\linewidth]{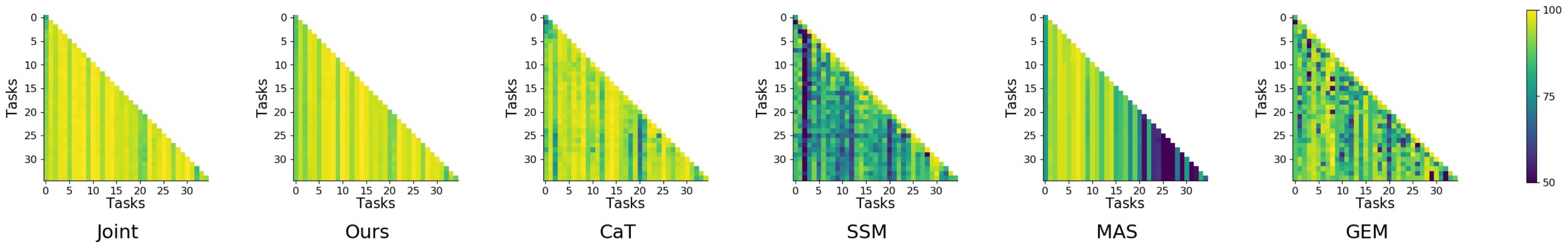}} \hfill
	\subfloat[Arxiv Dataset]{\includegraphics[width=\linewidth]{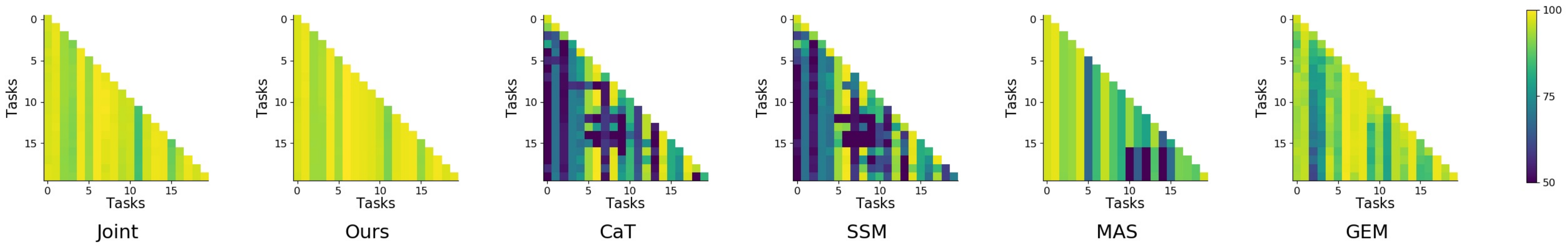}} \hfill
	\subfloat[Reddit Dataset]{\includegraphics[width=\linewidth]{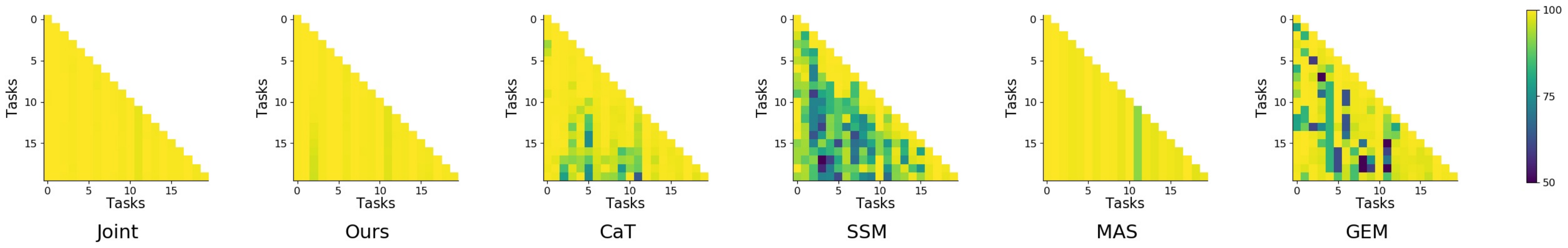}} \hfill
	\subfloat[Products Dataset]{\includegraphics[width=\linewidth]{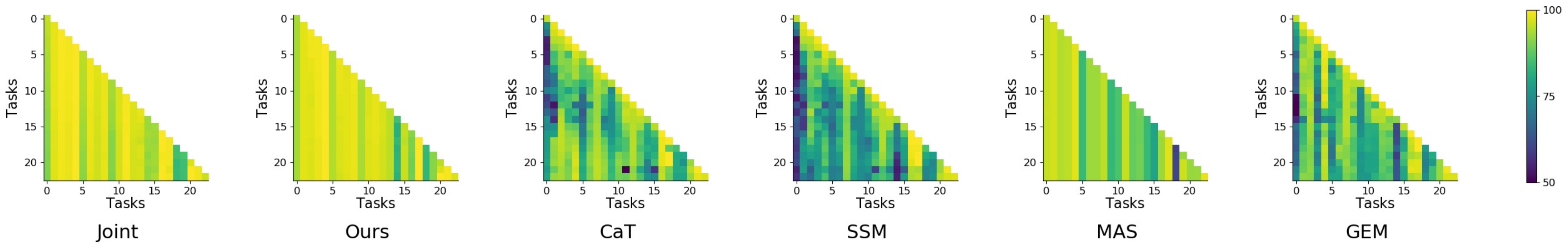}} \hfill
	\caption{Performance matrix visualization of Joint, Ours, CaT, SSM, MAS and GEM on CoraFull, Arxiv, Reddit and Products datasets (from top to bottom). Each entry in these matrices represents the performance of task $j$ (column) after learning task $i$ (row). Light colours indicate high accuracy and dark colours indicate low accuracy. Column $i$ from top to bottom can represent the change in the model's accuracy on all learned tasks after the model has learned task $\mathcal{T}_i$.}
	\label{fig_performance_matrix}
\end{figure*}

\subsubsection{Visualization of Performance Matrices}

To gain a deeper insight into the changes in model performance and forgetting ratio in dynamic CGL scenarios, we visualize the performance matrices of our method and several baselines across four datasets, under the memory consumption setting of three prompts, in Fig.~\ref{fig_performance_matrix}. Each triangular matrix represents the performance of a model within continuous learning setting. Specifically, each entry in these matrices denotes the performance of task $j$ (column) after the model learning task $i$ (row). The color gradient within each matrix represents the accuracy of the model, with lighter colors representing higher accuracy and darker colors representing lower accuracy. 

It is evident that most models exhibit strong performance on newly introduced tasks, as indicated by the yellow diagonal. However, regularization-based methods, such as MAS, demonstrate diminished long-term learning capabilities, particularly noticeable in the CoraFull dataset where the last few columns darken. As new tasks are introduced, many models suffer from catastrophic forgetting, which is reflected by the darkening below the diagonal. For example, CaT, SSM, and GEM show substantial declines in performance on earlier tasks (left columns) as they acquire new tasks (increasing rows) across all datasets. This underscores the critical challenge of mitigating catastrophic forgetting in CGL. In addition to the various CGL methods, we include the performance matrix of the Joint model, which serves as the theoretical upper bound for CGL tasks when solely training GNNs. The Joint model exhibits minimal performance degradation since it retains access to all historical data throughout the learning process, resulting in consistently high accuracy across all columns. 

Comparatively, the performance matrix of \ourmethod closely resembles that of the Joint model, indicating that \ourmethod effectively balances the acquisition of new tasks with the preservation of previously learned knowledge. This similarity highlights \ourmethod's robustness as a solution for CGL. Furthermore, on the Arxiv dataset, our method's performance matrix appears slightly brighter than that of the Joint model. This observation suggests that our prompt-based approach not only matches but can also enhance the theoretical performance upper bound, demonstrating its efficacy in improving continual learning outcomes.


\subsection{Diagnostic Experiments}

\subsubsection{Key Component Analysis} 

\ourmethod consists of two key components: \textit{Personalized Prompt Generator} (PG) and \textit{Hierarchical Prompting} (HP). Moreover, HP is decomposed of \textit{node-level prompts} (NP) and \textit{subgraph-level prompts} (SP). To investigate their efficacy, we evaluate various \ourmethod variants and report the AP and AF in TABLE~\ref{component_analysis}. According to TABLE~\ref{component_analysis}, we draw the following conclusions:

\begin{table}[t]
	\centering
	\caption{Key component analysis of the \ourmethod.}
	\hspace{-1.5em}
	\tablestyle{0.5}{4}{1.1}{
		\begin{tabular}{ccc||cc|cc}
			\thickhline
			\rowcolor{mygray}
			&  & & \multicolumn{2}{c|}{CoraFull} & \multicolumn{2}{c}{Arxiv} \\
			\rowcolor{mygray}
			\multicolumn{1}{c}{\multirow{-2}{*}{\textbf{PG}}}  & \multicolumn{1}{c}{\multirow{-2}{*}{\textbf{NP}}}  & \multicolumn{1}{c||}{\multirow{-2}{*}{\textbf{SP}}}  & AP(\%)  & AF(\%)  & AP(\%) & AF(\%) \\
			\hline \hline
			\XSolidBrush & \XSolidBrush & \XSolidBrush & 61.4±2.3 & -33.1±3.5 & 74.4±1.6 & -6.8±3.4 \\
			\XSolidBrush & \CheckmarkBold & \CheckmarkBold & 65.0±1.5 & -3.5±1.4 & 78.3±1.0 & -3.9±1.1 \\
			\CheckmarkBold & \XSolidBrush & \CheckmarkBold & 68.7±0.9 & 0.0±1.1 & 89.8±2.3 & -0.4±0.3 \\
			\CheckmarkBold & \CheckmarkBold & \XSolidBrush & 91.7±3.0 & 0.1±1.0 & 93.9±0.5 & -0.4±0.5 \\ 
			\hline
			\CheckmarkBold & \CheckmarkBold & \CheckmarkBold & 95.4±0.6 & -0.3±0.2 & 96.7±0.1 & -0.1±0.1 \\ 
			\hline
		\end{tabular}
	}
	\label{component_analysis}
\end{table}

\begin{table}[t]
	\centering
	\caption{Effect of prompt number on performance.}
	\resizebox{0.49\textwidth}{!}{
		\setlength\tabcolsep{6pt}
		\renewcommand\arraystretch{1.1}
		\begin{tabular}{c||cc|cc}
			\thickhline
			\rowcolor{mygray}
			& \multicolumn{2}{c|}{CoraFull} & \multicolumn{2}{c}{Arxiv} \\
			\rowcolor{mygray}
			\multicolumn{1}{c||}{\multirow{-2}{*}{\textbf{Number}}} & AP(\%) & AF(\%) & AP(\%) & AF(\%) \\
			\hline \hline
			1  & 65.0±1.5 & -3.5±1.4 & 78.3±1.0 & -3.9±1.1 \\
			2  & 94.8±0.7 & -0.9±0.6 & 96.3±0.6 & -0.5±0.5 \\
			3  & 95.4±0.6 & -0.3±0.2 & 96.7±0.1 & -0.1±0.1 \\
			4  & 94.9±1.4 & -0.3±0.5 & 95.8±1.6 & -0.6±0.9 \\
			\hline
		\end{tabular}
	}
	\label{num_prompt}
\end{table}

(i) The removal of any individual component results in a significant decrease in AP and AF for \ourmethod. This underscores the essential role each component plays in the overall effectiveness of the model. At the same time, when all prompt-related components are removed, AF increases sharply, which proves the effectiveness of our graph-prompt-based approach in overcoming catastrophic forgetting in CGL. Specifically, when both PG and all prompt-related components are excluded, the model exhibits substantially higher AF, highlighting the efficacy of our prompt-based approach in mitigating forgetting. 

(ii) The ablation of either NP or SP alone leads to noticeable performance degradation, emphasizing the importance of both feature-level and topological information in CGL. Notably, removing NP causes the most pronounced decline in performance, with the AP on the CoraFull dataset dropping by 26.7\%. This significant reduction demonstrates that differences in node classes across task graphs are critical for the model's discriminative capability. Similarly, the exclusion of SP results in decreased performance, highlighting that structural variations between task graphs are vital for maintaining high accuracy. This aspect addresses a gap in previous graph prompt methods, which did not adequately account for structural differences.

(iii) Eliminating the PG component from \ourmethod leads to severe reductions in both AP and AF across all datasets, underscoring the necessity of providing personalized prompts for each node. Specifically, without PG, the AP on the CoraFull dataset decreases from 95.4\% to 65.0\%, and the AF worsens from -0.3\% to -3.5\%. On the Arxiv dataset, AP drops from 96.7\% to 78.3\%, and AF increases from -0.1\% to -3.9\%. These significant performance declines illustrate that the Personalized Prompts Generator is crucial for generating tailored prompts that enable the model to effectively leverage each node's unique features and topological context.

\subsubsection{Number of Prompts} 

TABLE~\ref{num_prompt} examines how the number of prompts affects the performance of \ourmethod. The results indicate that using two or three prompts achieves the highest AP and the lowest AF on both the CoraFull and Arxiv datasets. Specifically, two prompts result in AP scores of 94.8\% and 96.3\%, while three prompts slightly improve these to 95.4\% and 96.7\%. In contrast, increasing the number of prompts to four does not lead to further performance gains and may even cause a minor decrease in AP.

Furthermore, the analysis highlights the efficiency of the PG component. Utilizing only two or three prompts balances high accuracy with low memory consumption, avoiding the unnecessary overhead associated with more prompts. Additionally, employing a single prompt significantly degrades performance, demonstrating that multiple prompts are essential for effectively capturing the diverse features and topological contexts of different nodes. These findings confirm that an optimal number of prompts ensures both robust performance and memory efficiency in \ourmethod.

\begin{table}[t]
	\centering
	\caption{Effect of dimensions of prompts on performance.}
	\resizebox{0.49\textwidth}{!}{
		\setlength\tabcolsep{6pt}
		\renewcommand\arraystretch{1.1}
		\begin{tabular}{c||cc|cc}
			\hline\thickhline
			\rowcolor{mygray}
			& \multicolumn{2}{c|}{CoraFull} & \multicolumn{2}{c}{Arxiv} \\
			\rowcolor{mygray}
			\multicolumn{1}{c||}{\multirow{-2}{*}{\textbf{Dimension}}} & AP(\%) & AF(\%) & AP(\%) & AF(\%) \\
			\hline \hline
			32  & 95.4±0.6 & -0.3±0.2 & 96.7±0.1 & -0.1±0.1 \\
			64  & 94.3±1.8 & -1.1±1.1 & 96.4±0.4 & -0.3±0.4 \\
			128 & 95.2±1.0 & -0.5±1.2 & 96.5±0.1 & -0.3±0.1 \\
			256 & 95.2±0.7 & -0.8±0.8 & 96.3±0.2 & -0.3±0.2 \\
			\hline
		\end{tabular}
	}
	\label{dim_prompt}
\end{table}

\subsubsection{Dimensions of Prompts}

To evaluate the impact of prompt embedding dimensions on \ourmethod, we conducted experiments using dimensions of 32, 64, 128, and 256, as shown in TABLE~\ref{dim_prompt}. 

In general, higher-dimensional prompts can encapsulate more information, potentially enhancing model performance. However, the introduction of the PG enables low-dimensional prompts to provide sufficient information for accurate predictions. The results demonstrate that \ourmethod maintains high AP and low AF across all tested dimensions, with 32-dimensional prompts achieving performance comparable to higher dimensions. Additionally, since subgraph-level prompts are directly tied to their embedding dimensions, higher dimensions lead to increased memory consumption. This highlights the efficiency of using lower-dimensional prompts in \ourmethod, ensuring robust performance while minimizing memory usage.

\begin{figure*}[t!]
\centering
\subfloat[CoraFull dataset]{
	\includegraphics[width=0.23\linewidth]{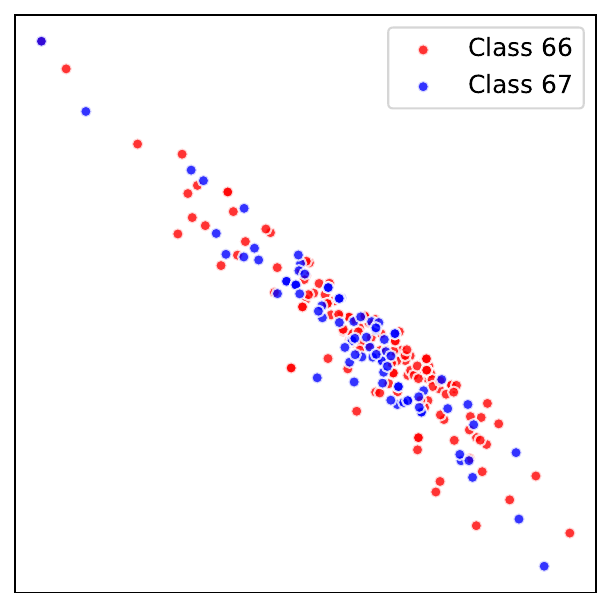} \hfill
	\includegraphics[width=0.23\linewidth]{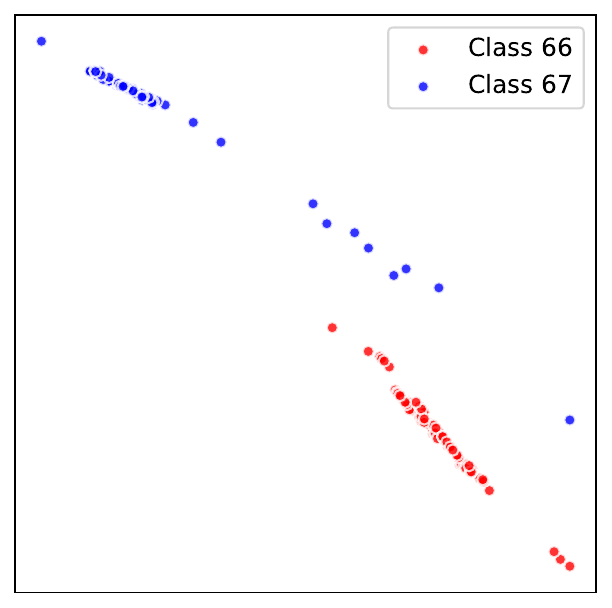}
} \hfill
\subfloat[Arxiv dataset]{
	\includegraphics[width=0.23\linewidth]{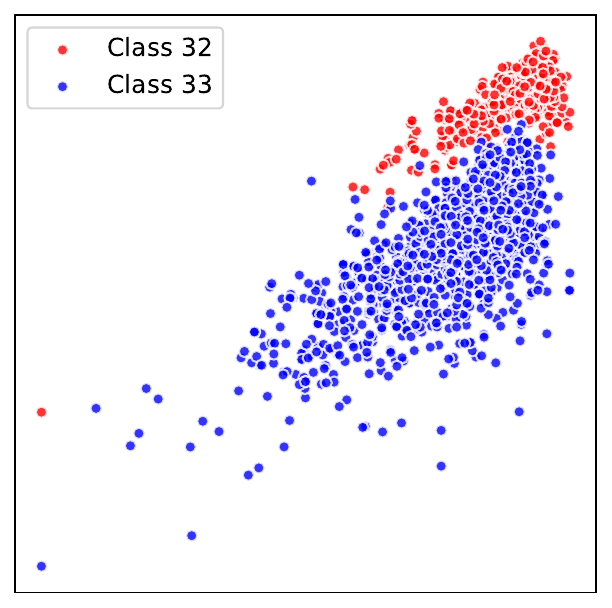} \hfill
	\includegraphics[width=0.23\linewidth]{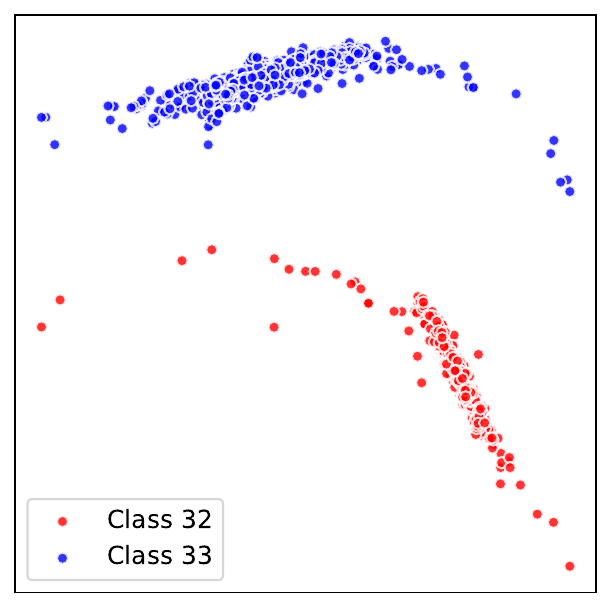}
} \hfill
\subfloat[Reddit dataset]{
	\includegraphics[width=0.23\linewidth]{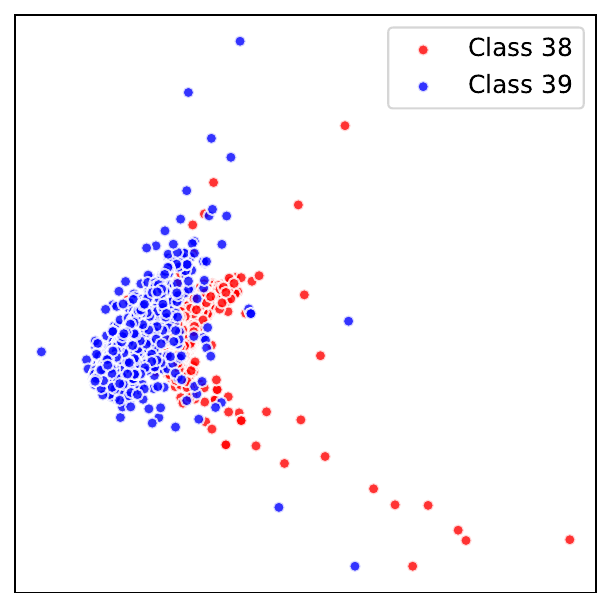} \hfill
	\includegraphics[width=0.23\linewidth]{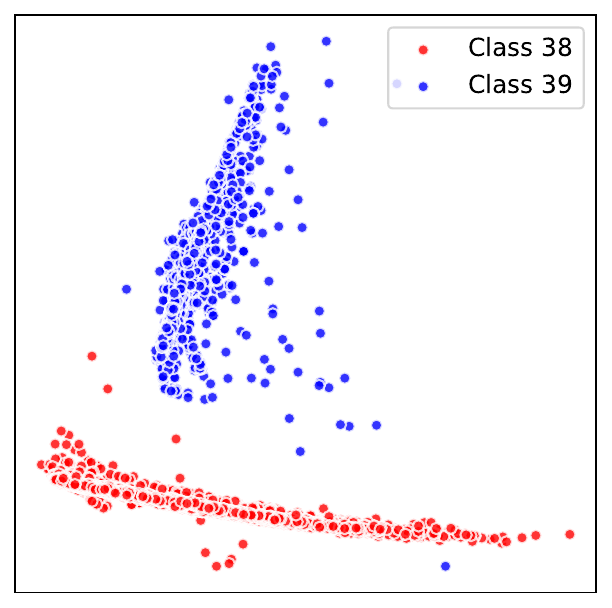}
} \hfill
\subfloat[Products dataset]{
	\includegraphics[width=0.23\linewidth]{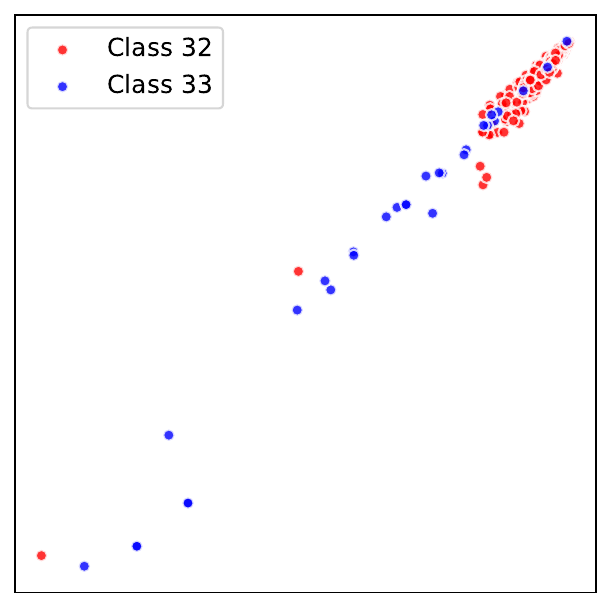} \hfill
	\includegraphics[width=0.23\linewidth]{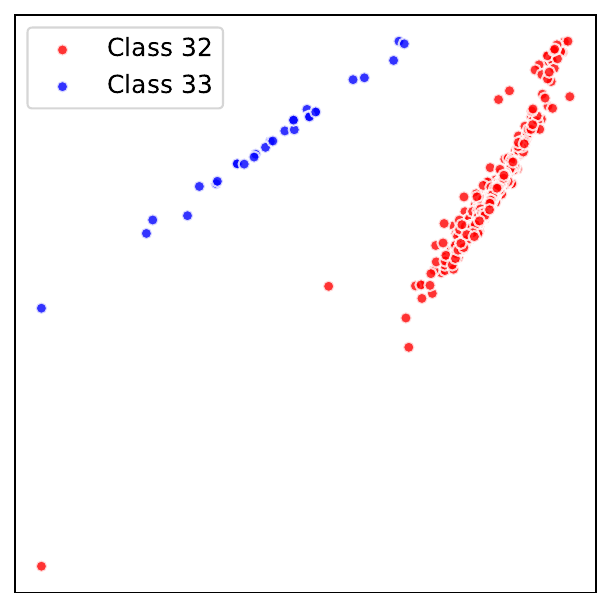}
} \hfill
\caption{The visualization of node embedding learned without (left) and with prompts (right) on four datasets.}
\label{visualization_embedding}
\end{figure*}

\begin{table}[t!]
\centering
\caption{Performance of different GNNs as backbone.}
\hspace{-1.5em}
\tablestyle{0.5}{1 }{1.1}{
	\begin{tabular}{c||cc|cc}
		\hline\thickhline
		\rowcolor{mygray}
		& \multicolumn{2}{c|}{CoraFull} & \multicolumn{2}{c}{Arxiv} \\
		\rowcolor{mygray}
		\multicolumn{1}{c||}{\multirow{-2}{*}{\textbf{Variant}}} & AP(\%) & AF(\%) & AP(\%) & AF(\%) \\
		\hline \hline
		\ourmethod(SAGE) & 94.4±0.5 & -0.6±0.3 & 95.9±0.4 & -0.8±0.4 \\
		\ourmethod(GAT)  & 93.0±0.4 & -1.0±0.8 & 95.1±1.1 & -1.4±0.9 \\
		\ourmethod(GCN)  & 95.4±0.6 & -0.3±0.2 & 96.7±0.1 & -0.1±0.1 \\
		\hline
	\end{tabular}
}
\label{diff_GNN_backbone}
\end{table}

\subsubsection{Robustness to Popular GNNs} 

To assess the applicability and robustness of \ourmethod across different GNNs, we instantiated the backbone of our method with three popular GNN architectures: GCN~\cite{kipf2016semi}, GAT~\cite{velivckovic2017graph}, and SAGE~\cite{hamilton2017inductive}. This resulted in three distinct variants of our approach, each utilizing one of these well-established GNN backbones, as summarized in TABLE~\ref{diff_GNN_backbone}. 

The results show that all three variants achieve SOTA accuracy in terms of AP while maintaining extremely low AF across the different datasets. These findings demonstrate that \ourmethod is highly adaptable, effectively integrating with a diverse range of GNN architectures without requiring any modifications to the underlying models. This flexibility underscores the versatility of our approach, enabling it to leverage the strengths of various GNNs while consistently mitigating catastrophic forgetting and ensuring robust performance across different graph learning scenarios.

\begin{figure}[t!]
	\centering
	\subfloat{\includegraphics[width=0.49\linewidth]{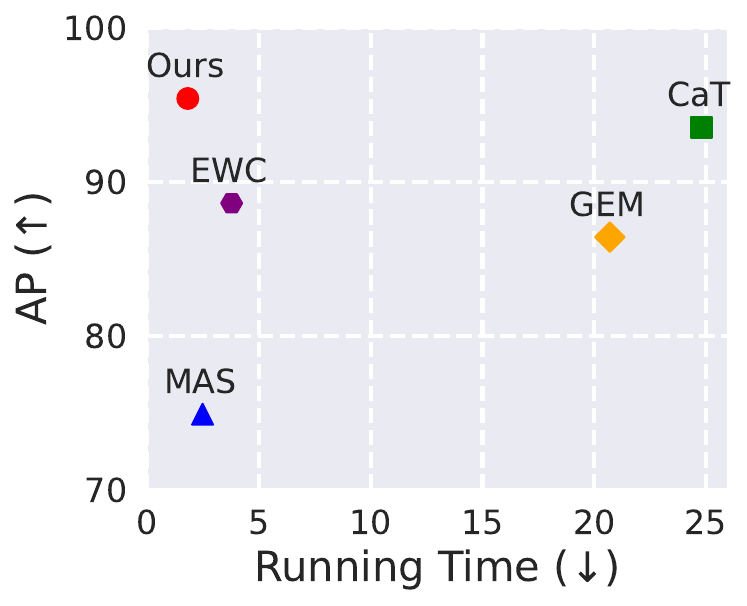}}
	\hfill
	\subfloat{\includegraphics[width=0.49\linewidth]{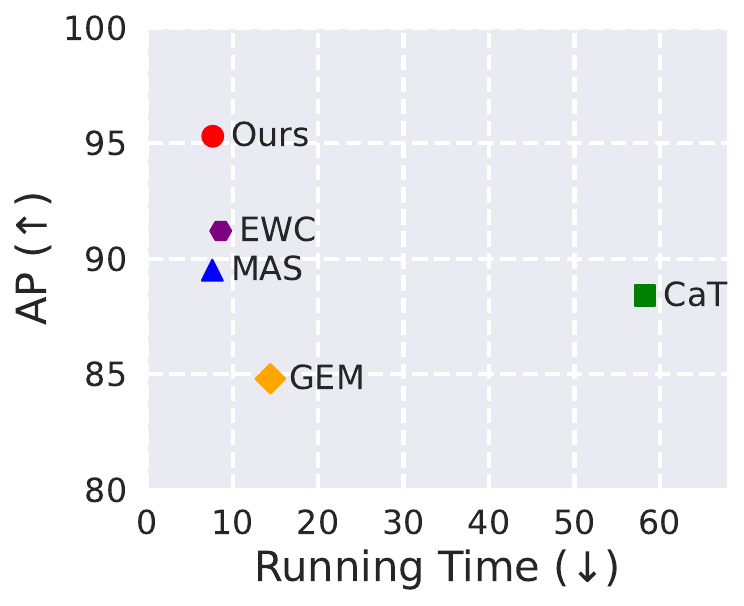}}
	\caption{Comparison of running time and AP on CoraFull (left) and Products (right) dataset.}
	\label{runtime_AP}
\end{figure}

\subsubsection{Running Time}

We evaluated the runtime and AP of several representative methods on the CoraFull and Products datasets. As shown in Fig.~\ref{runtime_AP}, \ourmethod achieves the highest AP with a relatively short runtime on both datasets, demonstrating its efficiency in CGL. 

This highlights the effectiveness of our parameter-efficient prompt fine-tuning strategy. According to TABLE~\ref{main_result}, replay-based methods, such as CaT, achieve near-SOTA AP but exhibit significantly higher runtimes as graph size increases, primarily due to the computational overhead of graph condensation and retraining data in memory. While effective for smaller graphs, this makes such approaches less suitable for large-scale applications. Regularization-based methods, like EWC and MAS, have lower runtimes but fail to deliver competitive AP, as their reliance on preserving prior task knowledge limits adaptability to new tasks. Overall, these results showcase \ourmethod's ability to achieve state-of-the-art performance with reduced runtime, offering a scalable and efficient solution for CGL scenarios.

\subsubsection{Visualization of Node Embedding} 

To demonstrate the impact of our prompt method on node embeddings, we provided examples of model outputs with and without prompts for each dataset, as shown in Fig.~\ref{visualization_embedding}. Each visualization corresponds to a specific task within a dataset, where the embeddings are projected into two dimensions using Principal Component Analysis (PCA) for clarity.

Without prompts, embeddings of nodes from different classes within the same task tend to overlap significantly, as seen in the left panels in each dataset of Fig.~\ref{visualization_embedding}. This lack of separation hinders the model's ability to distinguish between classes effectively. In contrast, with the incorporation of prompts, embeddings become more distinct and well-separated, as shown in the right panels of Fig.~\ref{visualization_embedding}. This highlights the role of prompts in enhancing the model's ability to leverage feature and topological information for better class separation. These results clearly demonstrate the effectiveness of our prompt-based approach in improving the discriminative power of node embeddings, thereby boosting the overall performance of \ourmethod in CGL scenarios.

\begin{figure}[t]
	\subfloat{\includegraphics[width=0.5\linewidth]{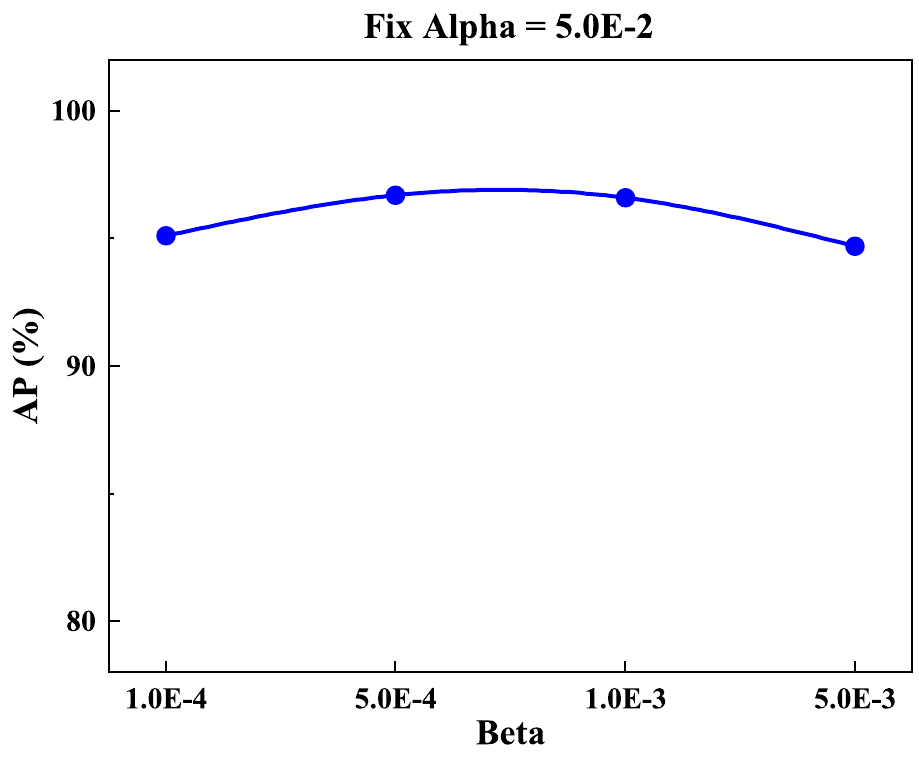}} \hfill
	\subfloat{\includegraphics[width=0.49\linewidth]{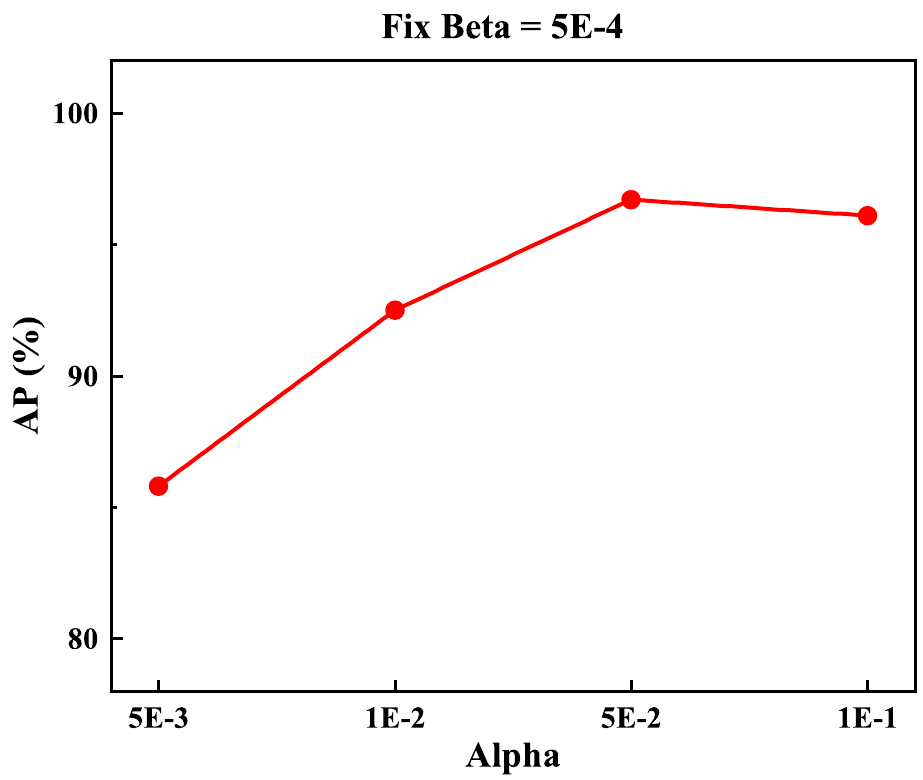}}
	\caption{Comparison of different learning rates of prediction layer (left) and prompts (right).}
	\label{fig7}
\end{figure}

\subsubsection{Hyperparametric analysis}

Here we study the effect of the learning rate $\alpha$ of the prompts and the learning rate $\beta$ of the prediction layer on AP. We performed a grid search over their values on the Arxiv dataset. First, we set $\alpha = 0.05$ and varied $\beta$, then set $\beta = 5\times10^{-4}$ and varied $\alpha$. As shown in Fig.~\ref{fig7}, the AP is largely unaffected by changes in $\beta$, as illustrated in Fig.~\ref{fig7} (left), indicating that our method is robust to the learning rate of the prediction layer. On the other hand, the AP decreases significantly when $\alpha$ is too small, as shown in Fig.~\ref{fig7} (right), highlighting the critical role of prompt tuning in the learning process. Since the model backbone remains frozen, the prompts play a vital role in adapting the model to new tasks, as evidenced by the improved performance when $\alpha$ is appropriately tuned.

\section{Conclusion}

This paper presents \ourmethod, a novel framework designed to tackle memory consumption and data privacy challenges in CGL. For the first time, \ourmethod incorporates graph prompt learning into CGL, employing hierarchical prompting to instruct the model through features and topologies to address the variability of task graphs in CGL. Our personalized prompt generator generates tailored prompts for each node while reducing spatial complexity from $O(N \cdot d)$ to $O(k \cdot d)$, demonstrating optimal performance with $k=3$. Extensive experiments show that our method achieves SOTA performance while effectively minimizing memory usage and safeguarding data privacy.





 
\vspace{11pt}

\vspace{11pt}

\vfill


\begin{thebibliography}{1}
\bibliographystyle{IEEEtran}

\bibitem{wu2020comprehensive}
Z.~Wu, S.~Pan, F.~Chen, G.~Long, C.~Zhang, and S.~Y. Philip, ``A comprehensive survey on graph neural networks,'' \textit{IEEE transactions on neural networks and learning systems}, vol.~32, no.~1, pp. 4--24, 2020.

\bibitem{yuan2011efficient}
Y.~Yuan, G.~Wang, H.~Wang, and L.~Chen, ``Efficient subgraph search over large uncertain graphs,'' \textit{Proceedings of the VLDB Endowment}, vol.~4, no.~11, pp. 876--886, 2011.

\bibitem{li2024cgmega}
H.~Li, Z.~Han, Y.~Sun, F.~Wang, P.~Hu, Y.~Gao, X.~Bai, S.~Peng, C.~Ren, X.~Xu \textit{et~al.}, ``Cgmega: explainable graph neural network framework with attention mechanisms for cancer gene module dissection,'' \textit{Nature Communications}, vol.~15, no.~1, p. 5997, 2024.

\bibitem{wang2022weighted}
Z.~Wang, Y.~Chai, C.~Sun, X.~Rui, H.~Mi, X.~Zhang, and S.~Y. Philip, ``A weighted symmetric graph embedding approach for link prediction in undirected graphs,'' \emph{IEEE Transactions on Cybernetics}, vol.~54, no.~2, pp. 1037--1047, 2022.

\bibitem{xie2022semisupervised}
Y.~Xie, Y.~Liang, M.~Gong, A.~K. Qin, Y.-S. Ong, and T.~He, ``Semisupervised graph neural networks for graph classification,'' \emph{IEEE Transactions on Cybernetics}, vol.~53, no.~10, pp. 6222--6235, 2022.

\bibitem{wang2024noise}
Q.~Wang, A.~Wu, Y.~Yuan, Y.~Wang, G.~Zhong, X.~Gao, and C.~Yang, ``Noise-resistant graph neural networks for session-based recommendation,'' in \textit{Proceedings APWeb and WAIM Joint International Conference on Web and Big Data}. Springer, 2024, pp. 144--160.

\bibitem{xu2024grakerformer}
L.~Xu, H.~Liu, X.~Yuan, E.~Chen, and Y.~Tang, ``Grakerformer: A transformer with graph kernel for unsupervised graph representation learning,'' \emph{IEEE Transactions on Cybernetics}, vol.~54, no.~12, pp. 7320--7332, 2024. 

\bibitem{liu2021overcoming}
H.~Liu, Y.~Yang, and X.~Wang, ``Overcoming catastrophic forgetting in graph neural networks,'' in \textit{Proceedings of the AAAI conference on artificial intelligence}, vol.~35, no.~10, 2021, pp. 8653--8661.

\bibitem{wang2022lifelong}
C.~Wang, Y.~Qiu, D.~Gao, and S.~Scherer, ``Lifelong graph learning,'' in \textit{Proceedings of the IEEE/CVF conference on computer vision and pattern recognition}, 2022, pp. 13\,719--13\,728.

\bibitem{zhang2022adaptive}
J.~Zhang, D.~Zhou, and M.~Chen, ``Adaptive cointegration analysis and modified rpca with continual learning ability for monitoring multimode nonstationary processes,'' \emph{IEEE Transactions on Cybernetics}, vol.~53, no.~8, pp. 4841--4854, 2022.

\bibitem{zhang2022cglb}
X.~Zhang, D.~Song, and D.~Tao, ``Cglb: Benchmark tasks for continual graph learning,'' \textit{Advances in Neural Information Processing Systems}, vol.~35, pp. 13\,006--13\,021, 2022.

\bibitem{rakaraddi2022reinforced}
A.~Rakaraddi, L.~Siew~Kei, M.~Pratama, and M.~De~Carvalho, ``Reinforced continual learning for graphs,'' in \textit{Proceedings of the 31st ACM International Conference on Information \& Knowledge Management}, 2022, pp. 1666--1674.

\bibitem{cai2022multimodal}
J.~Cai, X.~Wang, C.~Guan, Y.~Tang, J.~Xu, B.~Zhong, and W.~Zhu, ``Multimodal continual graph learning with neural architecture search,'' in \textit{Proceedings of the ACM Web Conference 2022}, 2022, pp. 1292--1300.

\bibitem{zhang2024continual}
X.~Zhang, D.~Song, and D.~Tao, ``Continual learning on graphs: Challenges, solutions, and opportunities,'' \textit{arXiv preprint arXiv:2402.11565}, 2024.

\bibitem{zhang2022hierarchical}
X.~Zhang, D.~Song, and D.~Tao, ``Hierarchical prototype networks for continual graph representation learning,'' \textit{IEEE Transactions on Pattern Analysis and Machine Intelligence}, vol.~45, no.~4, pp. 4622--4636, 2023.

\bibitem{zhang2023continual}
P.~Zhang, Y.~Yan, C.~Li, S.~Wang, X.~Xie, G.~Song, and S.~Kim, ``Continual learning on dynamic graphs via parameter isolation,'' in \textit{Proceedings of the 46th International ACM SIGIR Conference on Research and Development in Information Retrieval}, 2023, pp. 601--611.

\bibitem{zhou2021overcoming}
F.~Zhou and C.~Cao, ``Overcoming catastrophic forgetting in graph neural networks with experience replay,'' in \textit{Proceedings of the AAAI Conference on Artificial Intelligence}, vol.~35, no.~5, 2021, pp. 4714--4722.

\bibitem{zhang2022sparsified}
X.~Zhang, D.~Song, and D.~Tao, ``Sparsified subgraph memory for continual graph representation learning,'' in \textit{Proceeding of the IEEE International Conference on Data Mining}. 2022, pp. 1335--1340.

\bibitem{kim2022dygrain}
S.~Kim, S.~Yun, and J.~Kang, ``Dygrain: An incremental learning framework for dynamic graphs.'' in \textit{Proceedings of the Thirty-First International Joint Conference on Artificial Intelligence}, 2022, pp. 3157--3163.

\bibitem{liu2023cat}
Y.~Liu, R.~Qiu, and Z.~Huang, ``Cat: Balanced continual graph learning with graph condensation,'' in \textit{Proceeding of the IEEE International Conference on Data Mining}, 2023, pp. 1157--1162.

\bibitem{zhang2023ricci}
X.~Zhang, D.~Song, and D.~Tao, ``Ricci curvature-based graph sparsification for continual graph representation learning,'' \textit{IEEE Transactions on Neural Networks and Learning Systems}, vol. 35, no. 12, pp. 17398-17410, 2024

\bibitem{wu2022federated}
C.~Wu, F.~Wu, L.~Lyu, T.~Qi, Y.~Huang, and X.~Xie, ``A federated graph neural network framework for privacy-preserving personalization,'' \textit{Nature Communications}, vol.~13, no.~1, p. 3091, 2022.

\bibitem{xu2023making}
Z.~Xu, C.~Wang, M.~Qiu, F.~Luo, R.~Xu, S.~Huang, and J.~Huang, ``Making pre-trained language models end-to-end few-shot learners with contrastive prompt tuning,'' in \textit{Proceedings of the Sixteenth ACM International Conference on Web Search and Data Mining}, 2023, pp. 438--446.

\bibitem{wang2022learning}
Z.~Wang, Z.~Zhang, C.-Y. Lee, H.~Zhang, R.~Sun, X.~Ren, G.~Su, V.~Perot, J.~Dy, and T.~Pfister, ``Learning to prompt for continual learning,'' in \textit{Proceedings of the IEEE/CVF conference on computer vision and pattern recognition}, 2022, pp. 139--149.

\bibitem{liu2023pre}
P.~Liu, W.~Yuan, J.~Fu, Z.~Jiang, H.~Hayashi, and G.~Neubig, ``Pre-train, prompt, and predict: A systematic survey of prompting methods in natural language processing,'' \textit{ACM Computing Surveys}, vol.~55, no.~9, pp. 1--35, 2023.

\bibitem{li2021prefix}
X.~L. Li and P.~Liang, ``Prefix-tuning: Optimizing continuous prompts for generation,'' in \textit{Proceedings of the 59th Annual Meeting of the Association for Computational Linguistics and the 11th International Joint Conference on Natural Language Processing (Volume 1: Long Papers)}, 2021, pp. 4582--4597

\bibitem{fang2024universal}
T.~Fang, Y.~Zhang, Y.~Yang, C.~Wang, and L.~Chen, ``Universal prompt tuning for graph neural networks,'' \textit{Advances in Neural Information Processing Systems}, vol.~36, 2024.

\bibitem{sun2023all}
X.~Sun, H.~Cheng, J.~Li, B.~Liu, and J.~Guan, ``All in one: Multi-task prompting for graph neural networks,'' in \textit{Proceedings of the 29th ACM SIGKDD Conference on Knowledge Discovery and Data Mining}, 2023, pp. 2120--2131.

\bibitem{sun2023graph}
X.~Sun, J.~Zhang, X.~Wu, H.~Cheng, Y.~Xiong, and J.~Li, ``Graph prompt learning: A comprehensive survey and beyond,'' \textit{arXiv preprint arXiv:2311.16534}, 2023.

\bibitem{tan2023virtual}
Z.~Tan, R.~Guo, K.~Ding, and H.~Liu, ``Virtual node tuning for few-shot node classification,'' in \textit{Proceedings of the 29th ACM SIGKDD Conference on Knowledge Discovery and Data Mining}, 2023, pp. 2177--2188.

\bibitem{sun2022gppt}
M.~Sun, K.~Zhou, X.~He, Y.~Wang, and X.~Wang, ``Gppt: Graph pre-training and prompt tuning to generalize graph neural networks,'' in \textit{Proceedings of the 28th ACM SIGKDD Conference on Knowledge Discovery and Data Mining}, 2022, pp. 1717--1727.

\bibitem{schick2021s}
T.~Schick and H.~Sch{\"u}tze, ``It’s not just size that matters: Small language models are also few-shot learners,'' in \textit{Proceedings of the 2021 Conference of the North American Chapter of the Association for Computational Linguistics: Human Language Technologies}, 2021, pp. 2339--2352.

\bibitem{kipf2016semi}
T.~N. Kipf and M.~Welling, ``Semi-supervised classification with graph convolutional networks,'' \textit{arXiv preprint arXiv:1609.02907}, 2016.

\bibitem{zhou2022learning}
K.~Zhou, J.~Yang, C.~C. Loy, and Z.~Liu, ``Learning to prompt for vision-language models,'' \textit{International Journal of Computer Vision}, vol. 130, no.~9, pp. 2337--2348, 2022.

\bibitem{zhou2022conditional}
K.~Zhou, J.~Yang, C.~C. Loy, and Z.~Liu, ``Conditional prompt learning for vision-language models,'' in \textit{Proceedings of the IEEE/CVF conference on computer vision and pattern recognition}, 2022, pp. 16\,816--16\,825.

\bibitem{jia2022visual}
M.~Jia, L.~Tang, B.-C. Chen, C.~Cardie, S.~Belongie, B.~Hariharan, and S.-N. Lim, ``Visual prompt tuning,'' in \textit{Proceedings of the European Conference on Computer Vision}, 2022, pp. 709--727.

\bibitem{chen2022adaptformer}
S.~Chen, C.~Ge, Z.~Tong, J.~Wang, Y.~Song, J.~Wang, and P.~Luo, ``Adaptformer: Adapting vision transformers for scalable visual recognition,'' \textit{Advances in Neural Information Processing Systems}, vol.~35, pp. 16\,664--16\,678, 2022.

\bibitem{deng2023prompt}
C.~Deng, Q.~Chen, P.~Qin, D.~Chen, and Q.~Wu, ``Prompt switch: Efficient clip adaptation for text-video retrieval,'' in \emph{Proceedings of the IEEE/CVF International Conference on Computer Vision}, 2023, pp. 15\,648--15\,658.

\bibitem{hu2022promptcap}
Y.~Hu, H.~Hua, Z.~Yang, W.~Shi, N.~A. Smith, and J.~Luo, ``Promptcap: Prompt-guided task-aware image captioning,'' \emph{arXiv preprint arXiv:2211.09699}, 2022.

\bibitem{zheng2024exploring}
F.~Zheng, J.~Cao, W.~Yu, Z.~Chen, N.~Xiao, and Y.~Lu, ``Exploring low-resource medical image classification with weakly supervised prompt learning,'' \emph{Pattern Recognition}, vol. 149, p. 110250, 2024.

\bibitem{du2022learning}
Y.~Du, F.~Wei, Z.~Zhang, M.~Shi, Y.~Gao, and G.~Li, ``Learning to prompt for open-vocabulary object detection with vision-language model,'' in \emph{Proceedings of the IEEE/CVF Conference on Computer Vision and Pattern Recognition}, 2022, pp. 14\,084--14\,093.

\bibitem{li2024learning}
J.~Li, J.~Zhang, J.~Li, G.~Li, S.~Liu, L.~Lin, and G.~Li, ``Learning background prompts to discover implicit knowledge for open vocabulary object detection,'' in \emph{Proceedings of the IEEE/CVF Conference on Computer Vision and Pattern Recognition}, 2024, pp. 16\,678--16\,687.

\bibitem{luo2024vscode}
Z.~Luo, N.~Liu, W.~Zhao, X.~Yang, D.~Zhang, D.-P. Fan, F.~Khan, and J.~Han, ``Vscode: General visual salient and camouflaged object detection with 2d prompt learning,'' in \emph{Proceedings of the IEEE/CVF Conference on Computer Vision and Pattern Recognition}, 2024, pp. 17\,169--17\,180.

\bibitem{liu2023graphprompt}
Z.~Liu, X.~Yu, Y.~Fang, and X.~Zhang, ``Graphprompt: Unifying pre-training and downstream tasks for graph neural networks,'' in \textit{Proceedings of the ACM Web Conference 2023}, 2023, pp. 417--428.

\bibitem{mccallum2000automating}
A.~K. McCallum, K.~Nigam, J.~Rennie, and K.~Seymore, ``Automating the construction of internet portals with machine learning,'' \textit{Information Retrieval}, vol.~3, pp. 127--163, 2000.

\bibitem{hu2020open}
W.~Hu, M.~Fey, M.~Zitnik, Y.~Dong, H.~Ren, B.~Liu, M.~Catasta, and J.~Leskovec, ``Open graph benchmark: Datasets for machine learning on graphs,'' \textit{Advances in neural information processing systems}, vol.~33, pp. 22\,118--22\,133, 2020.

\bibitem{hamilton2017inductive}
W.~Hamilton, Z.~Ying, and J.~Leskovec, ``Inductive representation learning on large graphs,'' \textit{Advances in neural information processing systems}, vol.~30, 2017.

\bibitem{kirkpatrick2017overcoming}
J.~Kirkpatrick, R.~Pascanu, N.~Rabinowitz, J.~Veness, G.~Desjardins, A.~A. Rusu, K.~Milan, J.~Quan, T.~Ramalho, A.~Grabska-Barwinska \textit{et~al.}, ``Overcoming catastrophic forgetting in neural networks,'' \textit{Proceedings of the national academy of sciences}, vol. 114, no.~13, pp. 3521--3526, 2017.

\bibitem{aljundi2018memory}
R.~Aljundi, F.~Babiloni, M.~Elhoseiny, M.~Rohrbach, and T.~Tuytelaars, ``Memory aware synapses: Learning what (not) to forget,'' in \textit{Proceedings of the European conference on computer vision}, 2018, pp. 139--154.

\bibitem{lopez2017gradient}
D.~Lopez-Paz and M.~Ranzato, ``Gradient episodic memory for continual learning,'' \textit{Advances in neural information processing systems}, vol.~30, 2017.

\bibitem{li2017learning}
Z.~Li and D.~Hoiem, ``Learning without forgetting,'' \textit{IEEE transactions on pattern analysis and machine intelligence}, vol.~40, no.~12, pp. 2935--2947, 2017.

\bibitem{velivckovic2017graph}
P.~Veli{\v{c}}kovi{\'c}, G.~Cucurull, A.~Casanova, A.~Romero, P.~Lio, and Y.~Bengio, ``Graph attention networks,'' \textit{arXiv preprint arXiv:1710.10903}, 2017.


\end{thebibliography}
\end{document}